%% file: sports_agent_anonymization.tex
\documentclass[dvipsnames]{article}
\usepackage{colm2024_conference}

\usepackage{booktabs}
\usepackage{enumitem}
\usepackage{wrapfig}
\usepackage{algorithm}
\usepackage{algpseudocode}
\usepackage{graphicx}
\usepackage[misc]{ifsym}
\usepackage{multicol}
\usepackage{microtype}
\usepackage{amsmath}
\usepackage{colortbl}
\usepackage[utf8]{inputenc}
\usepackage[T1]{fontenc}
\definecolor{lightgray}{rgb}{0.9,0.9,0.9}
\usepackage{caption}
\usepackage{subcaption}
\usepackage{setspace}
\usepackage{url}
\usepackage{multirow}
\usepackage{tabularx}
\usepackage{blindtext}
\usepackage{pgfplots}
\pgfplotsset{compat=1.18}
\usepackage{tikz}
\usetikzlibrary{er,positioning,bayesnet}
\usepackage{makecell}
\usepackage{siunitx}
\usepackage{nicefrac}
\usepackage{listings}
\usepackage[raster,skins, most]{tcolorbox}
\usepackage{xltabular}
\usepackage{adjustbox}
\usepackage{xurl}
\usepackage{rotating}
\usepackage{pifont}
\usepackage[normalem]{ulem}

\usepackage{fontawesome}
\usepackage{amsmath}
\usepackage{amssymb}

\usepackage[table]{xcolor}

\useunder{\uline}{\ul}{}

\input{math_commands.tex}

\newcommand*\justify{%
  \fontdimen2\font=0.4em% interword space
  \fontdimen3\font=0.2em% interword stretch
  \fontdimen4\font=0.1em% interword shrink
  \fontdimen7\font=0.1em% extra space
  \hyphenchar\font=`\-% allowing hyphenation
}

\renewcommand{\texttt}[1]{%
  \begingroup
  \ttfamily
  \begingroup\lccode`~=`/\lowercase{\endgroup\def~}{/\discretionary{}{}{}}%
  \begingroup\lccode`~=`[\lowercase{\endgroup\def~}{[\discretionary{}{}{}}%
  \begingroup\lccode`~=`.\lowercase{\endgroup\def~}{.\discretionary{}{}{}}%
  \catcode`/=\active\catcode`[=\active\catcode`.=\active
  \justify\scantokens{#1\noexpand}%
  \endgroup
}

\usepackage{makecell}
\usetikzlibrary{tikzmark}
\makeatletter
\newcommand*\myfontsize{%
  \@setfontsize\myfontsize{7}{8}%
}
\makeatother

\definecolor{uclablue}{RGB}{159, 195, 224}

\definecolor{uclagold}{RGB}{255, 240, 180}

\definecolor{aliceblue}{RGB}{255, 238, 241}

\definecolor{cadmiumgreen}{rgb}{0.0, 0.42, 0.24}

\definecolor{myred}{rgb}{0.7, 0.3, 0.0}
\definecolor{myblue}{rgb}{0.2, 0.3, 0.6}
\definecolor{babygreen}{rgb}{0.85, 0.97, 0.85}

\makeatletter
\newcommand{\blfootnote}[1]{%
  \begingroup
  \renewcommand{\thefootnote}{}%
  \renewcommand{\@makefntext}[1]{##1}%
  \footnotetext{#1}%
  \addtocounter{footnote}{-1}%
  \endgroup
}
\makeatother

\definecolor{purple1}{RGB}{126, 107, 196}
\definecolor{purple2}{RGB}{199, 158, 207}
\definecolor{purple3}{RGB}{214, 200, 255}
\definecolor{purple4}{RGB}{254, 240, 255}

\definecolor{deepblue}{RGB}{48, 58, 82}

\definecolor{darkorange}{RGB}{255, 140, 0}
\definecolor{darkblue}{RGB}{84, 112, 198}
\definecolor{lightgreen}{RGB}{145, 204, 117}
\definecolor{lightyellow}{RGB}{250, 200, 88}
\definecolor{lightred}{RGB}{238, 102, 102}
\definecolor{lightblue}{RGB}{115, 192, 222}

\definecolor{darkred}{rgb}{0.55, 0.0, 0.0}
\definecolor{navy}{rgb}{0.0, 0.0, 0.55}
\definecolor{darkgreen}{rgb}{0.0, 0.39, 0.0}

\newtcolorbox{promptbox}{
  breakable,
  colback=blue!10!white,
  arc=5pt,
  boxrule=0.5pt,
  colframe=blue!50!violet,
  before upper={\normalsize},
  fontupper=\fontfamily{ptm}\selectfont,
}

\definecolor{deepPurple}{HTML}{330066}

\definecolor{uclablue_old}{rgb}{0.15, 0.45, 0.68}
\hypersetup{
    breaklinks,
    citecolor=qwenblue,
    linkcolor=qwenblue,
    urlcolor=qwenblue,
    colorlinks=true,
}

\newtcolorbox{mybox}[2][]
  {colback = black!5!white, colframe = black!75!black, fonttitle = \bfseries,
    colbacktitle = black!100!black, enhanced, before upper={\fontsize{8}{11}\obeyspaces\obeylines\selectfont}, fontupper=\selectfont,
    attach boxed title to top left={yshift=-2.2mm,xshift=4mm},
    title=#2,#1}

\newtcolorbox{highlightbox}[2][]
  {colback = qwenblue!8,           % 背景：蓝色系，很淡
   colframe = qwenblue,            % 边框：与 Abstract 一致
   colbacktitle = qwenblue,        % 标题区底色：与边框同色
   coltitle = white,               % 标题字体：白色
   fonttitle = \bfseries,
   enhanced,
   before upper={\fontsize{8}{11}\obeyspaces\obeylines\selectfont},
   fontupper=\selectfont,
   attach boxed title to top left={yshift=-2.2mm,xshift=4mm},
   title=#2,#1}

\title{%
  \parbox{0.92\textwidth}{%
    \centering
    Sports Agent: A Self-Evolving System for Conversational\\
    Data Analytics%
  }%
}

\author{%
\makebox[0pt][c]{%
\parbox{0.95\textwidth}{%
\centering
\large\bfseries
Chao Chen\textsuperscript{*},
Fuqi Lin\textsuperscript{*},
Ze Chen\textsuperscript{*},
Ziyan Liu\textsuperscript{*},
Chen Li\textsuperscript{*},
Haishu Jiang,
\\[0.25em]
Bo Xu,
Zihan Jiang,
Yue Wu,
Zheng Wang,
Kun Wang,
\\[0.25em]
Zhihang Fu\textsuperscript{\(\dagger\)},
Jieping Ye\textsuperscript{\(\dagger\)}
\\[0.65em]
\normalsize\normalfont
Alibaba Token Hub, Alibaba Group
\\[0.25em]
\texttt{\{ercong.cc, zhihang.fzh, yejieping.ye\}@alibaba-inc.com}
}}}

\newcommand{\ignore}[1]{}

\begin{document}

\maketitle

% \blfootnote{\noindent\footnotesize
% \textsuperscript{*}Equal contribution. \quad \textsuperscript{\dagger} Corresponding author.
% }

\blfootnote{\noindent\footnotesize
\textsuperscript{*}Equal contribution. \quad
\textsuperscript{\(\dagger\)}Corresponding author.
\\[0.25em]
We would like to thank the Qwen Team at Alibaba Token Hub (ATH), Alibaba Group,
for providing the computational resources and foundation models (Qwen) used in
this research.
}

\begin{abstract}
Conversational sports analytics requires agents to interpret user queries, coordinate heterogeneous resources, and deliver accurate, evidence-based responses with low latency. In production, changes in query distributions, data resources, and model behavior can degrade the performance of static agent configurations, while manual maintenance is often slow and costly. These challenges motivate controlled, evidence-driven evolution to sustain reliable performance in changing environments.

We present Sports Agent, a self-evolving framework for reliable conversational sports analytics. It integrates an Agentic SQL Model for structured-data reasoning, a Sports Agent Harness for planning and tool orchestration, and a unified Feedback Engine connecting online execution with offline evolution. The engine assesses responses and execution trajectories, verifies factual claims, diagnoses failures, and produces actionable evolution signals. Model evolution combines SQL-to-Question Scaling to broaden coverage with Graph-Guided Scaling to generate data targeting diagnosed errors. The synthesized questions then undergo trajectory distillation and quality filtering for iterative supervised fine-tuning, followed by a separate Post-RL stage. Harness evolution converts recurrent production issues into bounded prompt and skill updates, validated through independent audits, execution replay, and regression testing. Performance-tolerance checks prevent unacceptable performance degradation during harness simplification.

Model evolution yields a Qwen3.6-35B-A3B-based Agentic SQL Model that outperforms Qwen3.8-Max on question answering and analysis tasks over sports data and achieves approximately three times the inference efficiency. Across 30 successive harness-evolution rounds, the aggregate performance score increases by 10.8 points. In a live sports analytics product, the evolved version improves the factual correctness score by 10.5 points and the response richness by 158.1\%, while reducing average latency by 41.3\% relative to the non-evolved version.
\end{abstract}

\vfill

\clearpage
\makeatletter
\def\addcontentsline#1#2#3{%
  \addtocontents{#1}{\protect\contentsline{#2}{#3}{\thepage}{\@currentHref}}%
}
\makeatother
\setcounter{tocdepth}{3}
\tableofcontents
\clearpage

\input{Sports_Agent_Anonymization/Introduction}
\input{Sports_Agent_Anonymization/Overview}
\input{Sports_Agent_Anonymization/Model-loop}
\input{Sports_Agent_Anonymization/Harness-loop}

\input{Sports_Agent_Anonymization/AutoEval}
\input{Sports_Agent_Anonymization/Evaluation}
\input{Sports_Agent_Anonymization/RelatedWork}
\input{Sports_Agent_Anonymization/Discussion}
\input{Sports_Agent_Anonymization/Conclusion}

\clearpage
\bibliography{biblio}
\bibliographystyle{colm2024_conference}

\clearpage
\appendix
\section*{Appendix}
\addcontentsline{toc}{section}{Appendix}
\input{Sports_Agent_Anonymization/Appendix}
\end{document}

%% file: math_commands.tex
\usepackage{amsmath,amsfonts,bm}

\def\eqref#1{equation~\ref{#1}}
\def\1{\bm{1}}

\DeclareMathAlphabet{\mathsfit}{\encodingdefault}{\sfdefault}{m}{sl}
\SetMathAlphabet{\mathsfit}{bold}{\encodingdefault}{\sfdefault}{bx}{n}

%% file: Sports_Agent_Anonymization/Introduction.tex
\section{Introduction}
\label{sec:introduction}

\subsection{Conversational Sports Data Analytics}
Sports Agent targets factual retrieval and analytical reasoning over heterogeneous sports data. Unlike LLM-based chatbots that rely primarily on parametric knowledge~\citep{openai2025gpt5}, a production-grade sports agent must select appropriate data sources and specialized tools, retrieve evidence from the external environment, and ground its responses in the retrieved evidence.

Conversational sports analytics draws on diverse and evolving data sources, including structured databases, knowledge bases and real-time web search~\citep{xia2024language,zheman2026generation}. Structured databases support precise queries over schedules, player profiles, game records, and statistics; knowledge bases and FAQs provide rules, terminology, and historical context; web search delivers the latest news, injury reports, and roster updates ~\citep{hui2026interact,qin2025route}. Differences in data structure, update frequency, and access mechanisms require the system to select appropriate sources and tools for each request. Simple factual questions may require a single lookup, while statistical analysis demands a precise understanding of metric definitions and aggregation logic. Questions involving real-time information or multiple aspects of a topic often require coordinated tool use and evidence synthesis across sources. A sports agent must therefore support an end-to-end workflow encompassing intent understanding, tool selection, evidence verification, and response generation based on the retrieved evidence. Conversational sports analytics is also highly interactive. Users expect accurate, timely, and verifiable responses with low latency~\citep{martinez2025sportsql}. Although multi-step planning, iterative tool invocation, and cross-source evidence verification can enhance factual reliability, they introduce additional retrieval and inference overhead. Maintaining factual accuracy while minimizing response latency under evolving data and database schemas therefore remains a central challenge for production-grade sports agents.

Conversational sports analytics integrates information from multiple sources to support a broad range of applications, including professional sports analysis, live-event engagement, and sports information services~\citep{xu2026survey}. StatMuse translates natural-language questions into structured sports-statistics queries, returning textual answers alongside the underlying results~\citep{statmuse_asking_questions}. Wimbledon Match Chat integrates live match data with generative AI to support in-match statistical queries and performance analysis~\citep{ibm_wimbledon_ai_2025,baughman2025match}. AthleteGPT, deployed at the Paris 2024 Olympic Games, used retrieval-augmented generation to provide approximately 11,000 athletes with guidance on venues, rules, and event information~\citep{intel_genai_sports_2024}. Collectively, these applications illustrate the growing role of conversational sports analytics as an interface to official databases, live-event data, and domain knowledge. However, integrating these resources does not, by itself, ensure sustained performance as data sources, tool interfaces, and model capabilities evolve. Maintaining reliable operation under such changes remains an open challenge.

\subsection{Progressive Degradation of Static Agent Systems}

Most existing agent systems follow an ``offline development--deployment--manual maintenance'' paradigm ~\citep{zhai2025agentevolver}. During development, the foundation model, prompts, skills, tool descriptions, and routing policies are configured for a target task distribution and operating environment, with system performance validated through offline evaluation. The effectiveness of this approach depends on the deployment conditions remaining broadly consistent with those used during development and evaluation. In sports applications, however, evolving user requirements, data resources, and model capabilities can progressively widen the gap between static system configurations and operational demands~\citep{gao2025self_evolving_agents,liu2026path}.

We evaluate the proposed framework in a live sports analytics product, whose deployment context and system configuration are described in Section~\ref{sec:nba-chat}. The initial agent system used a largely static configuration, maintained through manual diagnosis and periodic refinement. Production observations indicate that this approach is susceptible to performance degradation as the agent environment evolves. Three factors are particularly salient.

\textbf{Query Distribution Shift.}
User interests shift substantially across the competition calendar. During the regular season and playoffs, queries primarily concern game results, player and team statistics, and standings; during the offseason, attention shifts toward trades, the draft, and roster changes. In the observed deployment, the share of database-oriented queries fell from approximately 63\% during the playoffs to 43\% in the offseason. This shift can reduce the effectiveness of planning and tool-routing policies tuned to the previous query distribution.

\textbf{Data Source and Schema Drift.}
The data environment continually evolves as new sources and metrics are introduced, schemas are revised, and field definitions change. Agents must adapt how they access and interpret data accordingly. During a six-month production period, three data sources were added and more than 100 fields were modified. Without timely adaptation, agents relying on fixed schema descriptions, example queries, and predefined access paths may reference invalid fields, use outdated query logic, or misinterpret retrieved results.

\textbf{Model Capability Misalignment.}
Parametric knowledge is updated more slowly than sports facts and can diverge from current player status, game statistics, and standings. Upgrading or replacing a model component can also change instruction following, planning, and tool-use behavior, affecting existing prompts and skills. Conversely, stronger models may render some harness components redundant. Model changes therefore require a reassessment of the surrounding harness configuration.

\begin{figure}[htbp]
  \centering
  \includegraphics[width=0.98\linewidth]{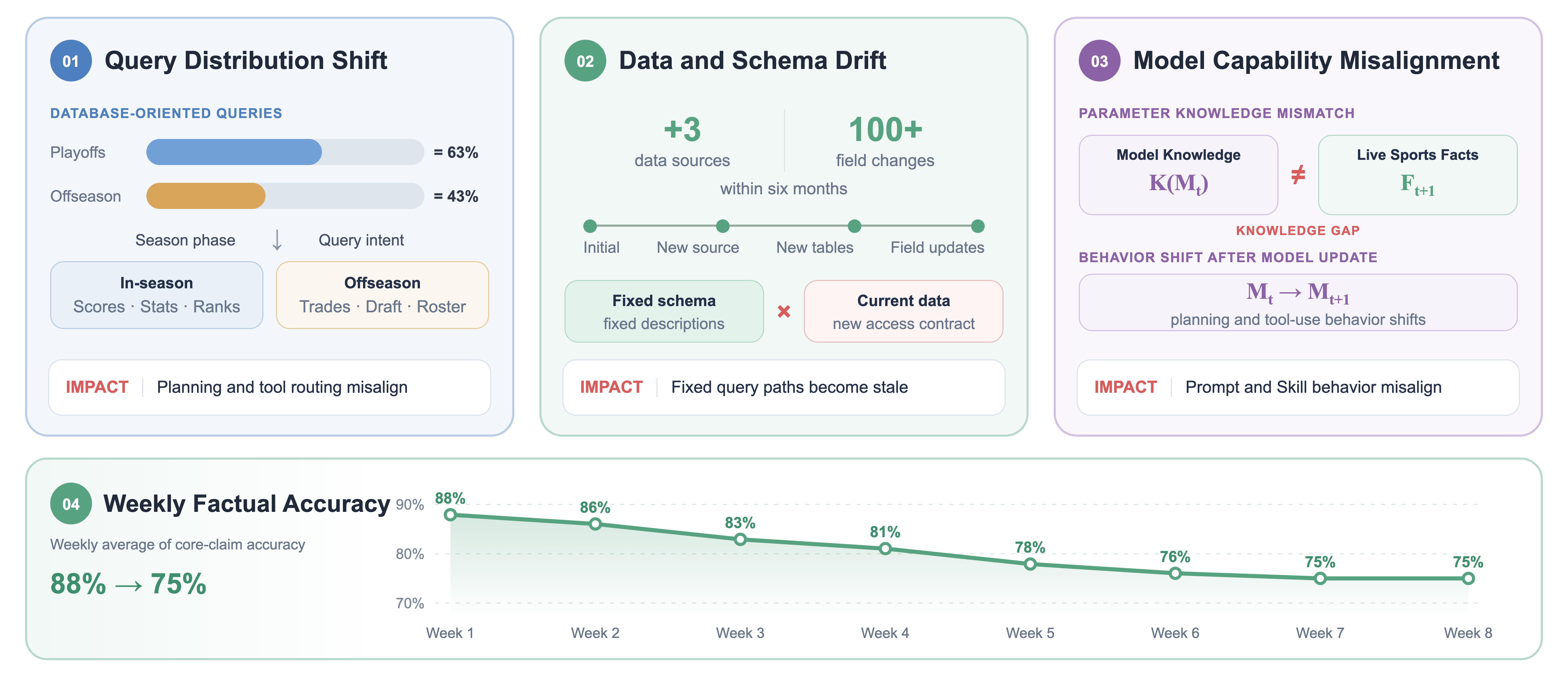}
  \caption{Progressive degradation of a static sports agent. The upper panels illustrate query shifts, data drift, and model-capability misalignment; the lower panel shows declining weekly factual accuracy.}
  \label{fig:degradation-of-static-agent}
\end{figure}

As illustrated in Figure~\ref{fig:degradation-of-static-agent}, these changes can progressively undermine planning, data access, and execution. In the observed production system, weekly factual accuracy declined from 88\% to 75\% over eight weeks. Manual diagnosis and ad hoc patches offer limited responsiveness and coverage, while accumulating prompts, rules, and skills can overfit individual cases and increase harness complexity. These limitations motivate continuous evolution guided by runtime evidence and the Feedback Engine.

\subsection{Research Problem and Scope}
These challenges motivate the research topic of this report: \textbf{How can runtime evidence guide the continuous evolution of both the model and the agent harness to maintain reliable performance in an evolving environment?}

We investigate three complementary aspects of this problem:

\textbf{Model Self-Evolution.} How can runtime failures and capability gaps inform the construction of high-quality training data, and how can iterative data refinement and post-training improve model generalization and efficiency?

\textbf{Harness Self-Evolution.} How can runtime evidence guide the diagnosis and targeted refinement of prompts and skills to improve task planning, tool use, and response style as task distributions evolve?

\textbf{Unified Feedback Engine.} How can the analysis of responses and execution trajectories support root-cause diagnosis and generate actionable feedback to guide both model and harness evolution?

\subsection{Contributions}
The main contributions of this report are as follows.

\textbf{A Dual-Loop Self-Evolution Framework.}
We introduce Sports Agent, which integrates the Sports Agent Harness, Agentic SQL Model, and Feedback Engine into a unified system. By connecting online task execution, runtime evidence collection, evaluation and diagnosis, and capability updates, the system establishes self-evolution loops for reliable conversational sports analytics.

\textbf{Data-Scaling-Driven Model Self-Evolution.}
We develop a data evolution pipeline encompassing question synthesis, trajectory distillation, error attribution, and targeted data generation. Combined with model fine-tuning and reinforcement learning, this pipeline continuously improves the accuracy, execution efficiency, and generalization of the Agentic SQL Model for sports data analysis.

\textbf{Harness Evolution with Dual Audit and Complexity Control.}
We enable controlled updates to prompts, skills, and response strategies by translating recurrent production issues into bounded edits. Each edit undergoes an independent static audit and paired execution replay, followed by joint replay for final selection. Non-inferiority checks guide harness simplification, while inactive components are deactivated rather than deleted.

\textbf{A Unified Feedback Engine for Both Evolution Loops.}
The engine represents answers, evidence, and execution trajectories as verifiable objects. Through evidence verification, step-level error attribution, and expert feedback, it identifies update targets and translates diagnostic findings into actionable signals for both model and harness evolution. It also provides unified regression evaluation and release control for candidate versions.

\textbf{Production Validation.}
We deploy and evaluate Sports Agent in a live sports analytics product. Relative to the baseline version, the evolved system improves the factual correctness score by 10.5 points and increases average response length, a proxy for response richness, by 158.1\%, while reducing average end-to-end latency by 41.3\%. Across 30 successive harness-evolution rounds, the aggregate performance score increases by 10.8 points. These results demonstrate improvements in response quality and execution efficiency in the evaluated production setting.

Beyond sports analytics, Sports Agent offers practical lessons for building self-evolving agents: accumulate evidence from multiple related failures before making persistent updates, identify shared root causes to generate training examples beyond the original cases, and use executable checks to validate generated training data. These principles can guide self-evolution in coding agents, enterprise assistants, and other agent systems. Section~\ref{sec:related-synthesis} discusses the design principles for self-evolving agents and the cross-domain transferability of our approach.

%% file: Sports_Agent_Anonymization/Overview.tex
\section{Sports Agent: Dual-Loop Evolution}
\label{sec:overview}
To address the challenges in Section~\ref{sec:introduction}, Sports Agent establishes two coordinated evolution loops for the Agentic SQL Model and the Sports Agent Harness. A unified Feedback Engine translates runtime evidence into diagnostic findings and update signals, providing evaluation and release control for candidate versions. This section formulates the evolution problem in a changing environment and presents the system architecture, covering online execution and offline evolution workflows.

\subsection{Problem Formulation}
\label{subsec:problem-formulation}

Let $\mathcal{E}_t$ denote the operating environment at time $t$, including the user-query distribution, data resources, and tool space $\mathcal{T}_t$. The distribution shift, data and schema drift, and model changes discussed in Section~\ref{sec:introduction} can be summarized as $\mathcal{E}_t \neq \mathcal{E}_{t+1}$. Consequently, an agent configuration that is effective in the current environment may become progressively less suitable as operating conditions change.

Given a user query $q_t$, the current Agentic SQL Model $M_t$ and Sports Agent Harness $H_t$ jointly execute the online task:

\begin{equation}
(y_t,\tau_t)
=
\operatorname{Execute}
\left(q_t;M_t,H_t,\mathcal{E}_t\right),
\label{eq:online-execution}
\end{equation}

where $y_t$ is the final answer and $\tau_t$ is the execution trajectory, including task planning, model inference, tool calls, tool results, and response generation. The system organizes the input query, final answer, and execution trajectory as runtime evidence,

\begin{equation}
e_t = \left(q_t,y_t,\tau_t\right),
\label{eq:runtime-evidence}
\end{equation}

Only task-relevant information from this representation is used for response quality assessment, failure attribution, and targeted capability updates. The use of this information follows the data privacy safeguards described in Appendix~\ref{app:data-privacy}.

Sports Agent optimizes two components: the Agentic SQL Model $M_t$, which supports sports-domain semantic understanding, complex SQL reasoning, and structured-data querying; and the Sports Agent Harness $H_t$, which governs planning, tool orchestration, prompts, skills, and response generation. The Feedback Engine converts runtime evidence into targeted evolution signals and guides validation and release decisions for candidate updates:

\begin{equation}
\begin{aligned}
(g_t^H,g_t^M)
  &= \operatorname{FeedbackEngine}(e_t),\\
H_{t+1}
  &= \mathcal{U}_H\left(H_t;g_t^H\right),\\
M_{t+1}
  &= \mathcal{U}_M\left(M_t;g_t^M\right).
\end{aligned}
\label{eq:coevolution-update}
\end{equation}

Here, $g_t^H$ and $g_t^M$ denote harness- and model-oriented evolution signals, respectively. The system aims to improve answer correctness and completeness while reducing end-to-end latency. Each candidate update undergoes independent evaluation and regression testing to identify potential trade-offs and guard against regressions in existing capabilities.

\subsection{Architecture Overview}
\label{subsec:architecture-overview}

\begin{figure}[htbp]
  \centering
  \includegraphics[width=\linewidth]{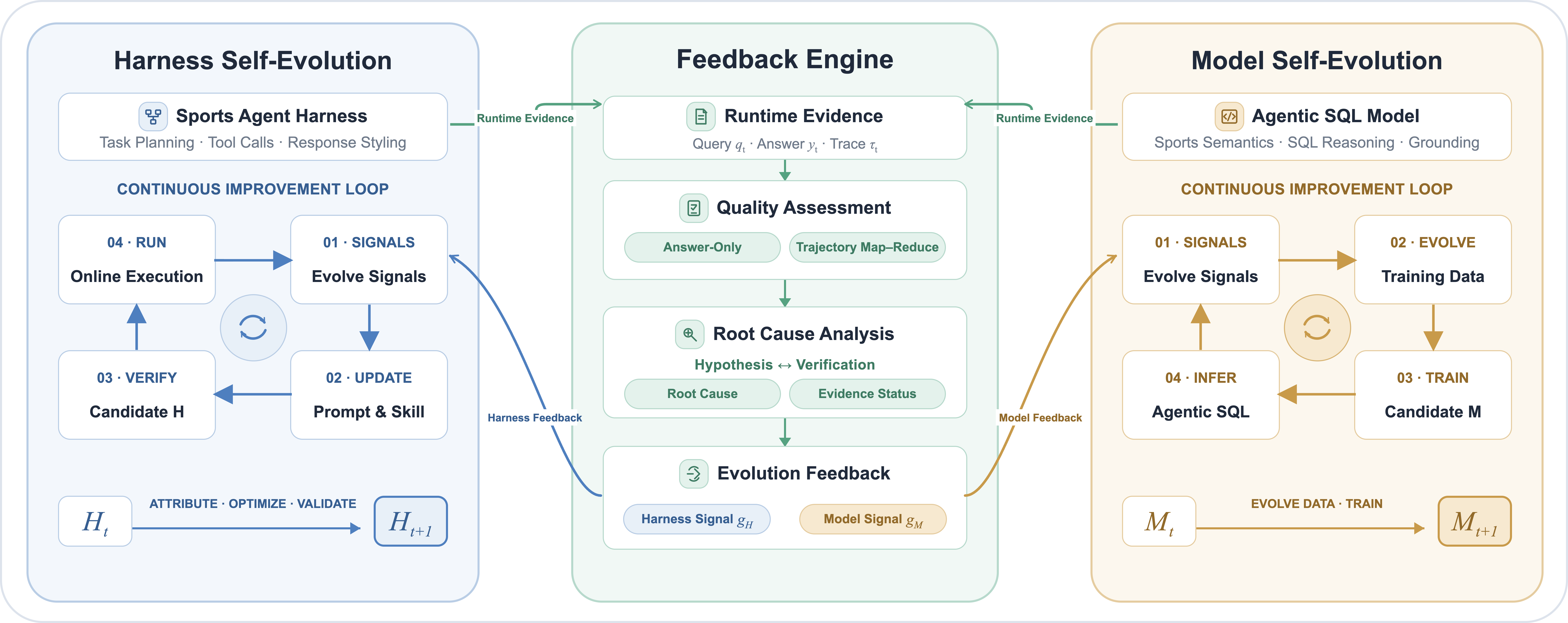}
  \caption{Sports Agent dual-loop evolution architecture. Feedback Engine converts runtime evidence into diagnostic findings and evolution signals that drive updates to the harness and model.}
  \label{fig:framework-overview}
\end{figure}

The three sources of degradation identified in Section~\ref{sec:introduction} require complementary forms of adaptation. Query shifts call for changes to planning, tool routing, prompts, and skills; data and schema changes require updated access strategies and model capabilities; and model upgrades may alter tool-use behavior or invalidate existing harness components. These challenges cannot be addressed through isolated model updates or incremental prompt engineering alone. Accordingly, Sports Agent comprises three tightly connected modules: \textbf{Harness Self-Evolution}, a \textbf{Unified Feedback Engine}, and \textbf{Model Self-Evolution}. As shown in Figure~\ref{fig:framework-overview}, the architecture organizes the harness and model evolution loops around the shared Feedback Engine. The harness loop adapts planning and tool orchestration to changes in tasks, tools, and model behavior, while the model loop addresses persistent capability gaps through targeted data construction and post-training. Positioned between the two loops, the Feedback Engine analyzes runtime evidence, diagnoses failures, routes target-specific evolution signals, and validates candidate updates.

\paragraph{Model Self-Evolution.}
This loop updates the specialized Agentic SQL Model; the parameters of the orchestration model responsible for task planning and tool selection are not updated in this work.  Structured databases provide the factual foundation for sports analytics, including schedules, game records, player statistics, and standings. The Agentic SQL Model provides the core capability for accessing and analyzing these data, making its accuracy and efficiency essential to overall system performance. Model-oriented diagnostic signals are compiled into targeted data-construction tasks. Question synthesis, trajectory distillation, and quality filtering then produce $M_{t+1}$ through in-loop supervised fine-tuning, improving sports semantics, complex SQL reasoning, cross-task generalization, and execution efficiency. A final reinforcement-learning stage outside the loop (Post-RL) produces $M_{\mathrm{final}}$.

\paragraph{Harness Self-Evolution.} 
This loop optimizes prompts and skills that guide the orchestration model’s task planning, tool-use policies, safety policies and response style. The Feedback Engine aggregates recurring failures into evidence-supported repair intents, guiding bounded edits to global rules, skill content, and activation conditions. Candidate edits undergo independent review, execution replay, and regression testing before a validated $H_{t+1}$  is released.

\paragraph{Unified Feedback Engine.}
Positioned between the two loops, the Feedback Engine performs the structured conversion from runtime evidence to system updates. It jointly assesses answers, supporting evidence, and execution trajectories; diagnoses root causes; routes evolution signals to the relevant component; and protects each update through regression validation and release gating.

The architecture connects online execution with offline evolution. During online execution, the orchestration model follows the harness configuration to plan tasks and coordinate tools, invoking the Agentic SQL Model when structured-data queries or analysis are required. Evidence retrieved from databases, web search, and knowledge bases is integrated into the final response. Relevant information derived from agent trajectories supports subsequent evaluation, failure diagnosis, and system evolution.

\begin{equation}
q_t
\longrightarrow H_t
\longrightarrow M_t
\longrightarrow \mathcal{T}_t
\longrightarrow \tau_t
\longrightarrow y_t.
\label{eq:online-path}
\end{equation}

Offline, the Feedback Engine assesses relevant execution evidence, diagnoses failures, routes model- and harness-oriented signals to their respective loops, and validates candidate versions:

\begin{equation}
e_t
\longrightarrow \operatorname{Feedback\ Engine}
\longrightarrow \left(g_t^H,g_t^M\right)
\longrightarrow \left(H_{t+1},M_{t+1}\right)
\longrightarrow \operatorname{Validate}.
\label{eq:offline-path}
\end{equation}

By closing the loop among online execution, automated assessment, and offline updates, Sports Agent continually recalibrates its capability boundary and maintains a more effective balance among answer quality, execution efficiency, and system complexity.

%% file: Sports_Agent_Anonymization/Model-loop.tex
\section{Model Self-Evolution Loop}
\label{sec:model-loop}
Model improvement requires not only more training data but also a training distribution that reflects observed failures and evolving task requirements. As discussed in Section~\ref{sec:introduction}, schema drift can render learned field references, metric definitions, and query logic outdated, while recurrent failures reveal capability gaps that data-volume expansion alone may not adequately address. We therefore organize task execution, failure diagnosis, data generation, and parameter updates into an iterative loop guided by execution feedback, with each training round targeting diagnosed capability gaps.

The model-evolution loop combines Feedback Engine diagnosis, targeted Data Scaling, and in-loop supervised fine-tuning (SFT). The Feedback Engine assesses runtime evidence and identifies model-side deficiencies (Section~\ref{sec:eval}), which guide SQL-to-Question Scaling and Graph-Guided Scaling. Questions generated by both methods undergo a shared trajectory-distillation and quality-filtering pipeline to produce training data for SFT. The updated model is then evaluated to inform the next evolution round. After iterative evolution concludes, a separate reinforcement-learning (RL) stage, termed Post-RL, further optimizes task performance.

Section~\ref{subsec:model-loop-task} formulates the Agentic SQL task and its execution trajectory. Section~\ref{subsec:model-loop-training} describes the evolution pipeline, the two Data Scaling methods, and in-loop SFT. Section~\ref{subsec:model-loop-post-rl} presents Post-RL, and Section~\ref{subsec:model-loop-experiments} evaluates model quality, execution efficiency, and the contribution of data evolution.

\subsection{Agentic SQL Task}
\label{subsec:model-loop-task}
Conversational sports data analytics draws on heterogeneous resources, but many user questions require precise computation over official databases. Answering them requires identifying the relevant entities, interpreting metric definitions, resolving temporal scope and implicit constraints, and verifying results against database evidence rather than relying on parametric knowledge. Conventional Text-to-SQL maps a natural-language question to a single SQL statement for a given schema and is typically evaluated through exact match or execution accuracy~\citep{li2023database,qin2025route}. In contrast, the Agentic SQL Model treats SQL as an intermediate action within a multi-step workflow: it explores the schema, constructs and executes queries, and refines them in response to execution feedback. Evaluation therefore considers both the execution trajectory and the final task outcome~\citep{huo2026bird}.

Given a user question, database resources, and an Agentic SQL tool space, the model solves a task through multi-step tool interaction. Let $q$ and $M$ denote the question and agentic model. The execution trajectory is
\begin{equation}
\tau=
\left[
(a_1,o_1),
(a_2,o_2),
\cdots,
(a_K,o_K)
\right].
\end{equation}

In the notation of Section~\ref{subsec:problem-formulation}, $\tau$ is the sub-trajectory that a single Agentic SQL call contributes to the system trajectory $\tau_t$ at run time.
Here, $K$ is the number of interaction steps and $(a_k,o_k)$ is the $k$-th tool-action--observation pair. The action $a_k$ may retrieve schema information, inspect tables or fields, execute SQL queries, or verify query results, while the observation $o_k$ captures the corresponding schema metadata, execution results, or error messages. The model selects the next action from the question and trajectory prefix:
\begin{equation}
a_k
\sim
\pi_M
\left(
a_k
\mid
q,\tau_{<k}
\right),
\end{equation}
where $\pi_M$ is the action policy induced by $M$ and $\tau_{<k}$ is the trajectory before step $k$. Once sufficient evidence has been collected or a termination condition is met, the system generates the final answer from the execution trajectory:
\begin{equation}
y=
\operatorname{Synthesize}
\left(q,\tau\right).
\end{equation}
Once invoked, the model autonomously explores the database schema, plans and executes queries, and uses execution feedback to correct syntax, field references, and query conditions. Relevant execution evidence, including query results and trajectory-level signals, supports subsequent evaluation, failure attribution, and data evolution.

The Agentic SQL Model is trained to improve query correctness and cross-database generalization while reducing interaction turns and end-to-end latency. To balance capability and inference cost, we use Qwen3.6-35B-A3B, a mixture-of-experts model with few active parameters, as the base model.

\subsection{Model Evolution Training}
\label{subsec:model-loop-training}
Model evolution requires both broader data coverage and targeted learning from observed failures. We formalize this process below.

Let $\mathcal{D}_0$ be the seed dataset curated from historical queries, and let $\Sigma$, $\mathcal{K}$, and $\mathcal{L}$ denote the database schema, sports-domain knowledge, and task-control variables, respectively. The first SQL-to-Question scaling round generates $\Delta\mathcal{D}_0$ under predefined controls and combines it with a filtered subset of $\mathcal{D}_0$:

\begin{equation}
\mathcal{D}_1=
\mathcal{D}_0^{\mathrm{sub}}
\cup
\Delta\mathcal{D}_0.
\end{equation}
Here, $\mathcal{D}_0$ is the seed training set, $\Delta\mathcal{D}_0$ is the generated increment in the first evolution round, and $\mathcal{D}_1$ is the expanded dataset. For each subsequent round $t\geq 1$, the Feedback Engine analyzes the agent execution trajectory, SQL results, judge outputs, and relevant schemas for the current model $M_t$, producing diagnostic findings $\mathcal{F}_t$. Each finding records the root cause, sports concept, semantic intent, related schema, and error pattern of a failed case. $\mathcal{F}_t$ is diagnostic metadata rather than a training sample or a complete synthesis specification.

The data increment for round $t$ is the union of SQL-to-Question and Graph-Guided Scaling:
\begin{equation}
\Delta\mathcal{D}_t=
\Delta\mathcal{D}_t^{\mathrm{SQL}}
\cup
\Delta\mathcal{D}_t^{\mathrm{Graph}}.
\end{equation}
The combined increment updates the training set and current model:
\begin{equation}
\mathcal{D}_{t+1}=
\mathcal{D}_t^{\mathrm{sub}}
\cup
\Delta\mathcal{D}_t,
\qquad
M_{t+1}=
\operatorname{SFT}(M_t,\mathcal{D}_{t+1}).
\end{equation}
The updated $M_{t+1}$ is evaluated by the Feedback Engine in the next round, coupling changes in the data distribution to changes in model capability.

\begin{figure}[htbp]
  \centering
  \includegraphics[width=\linewidth]{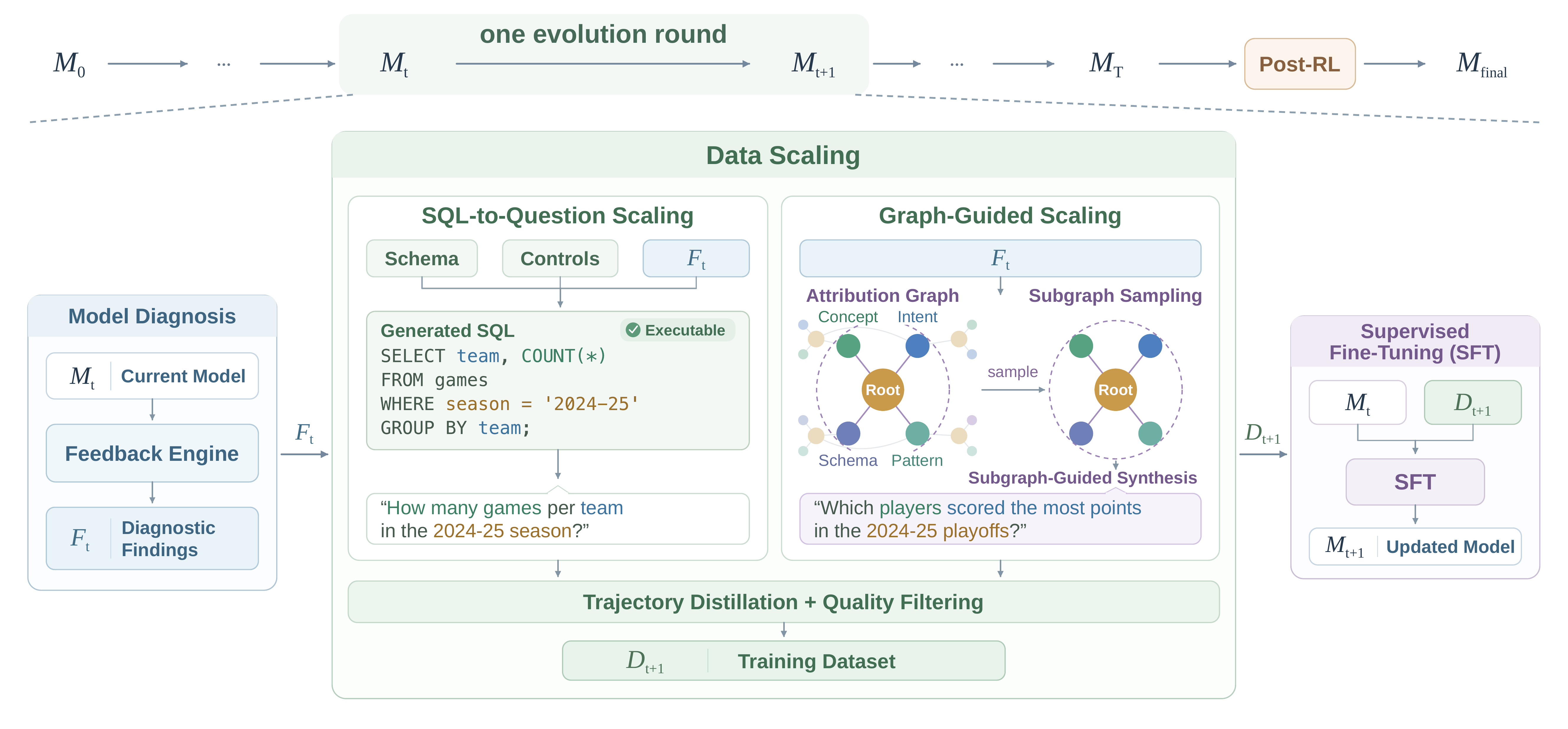}
  \caption{Model self-evolution loop. The Feedback Engine produces diagnostic findings $\mathcal{F}_t$ for SQL-to-Question and Graph-Guided Scaling; the resulting data update drives SFT in each round.}
  \label{fig:model-self-evolution-loop}
\end{figure}

\subsubsection{SQL-to-Question Scaling}
We propose a reverse SQL-to-Question synthesis method that uses executable SQL as an intermediate semantic representation. The pipeline first constructs the query logic and then expresses it as a natural-language question. This allows query structure, task difficulty, and domain semantics to be explicitly constrained at the SQL level, while execution verifies answerability. Compared with generating questions directly from a schema, this approach checks constraints before question generation, helping preserve constraint completeness and semantic consistency.

% For sample $i$, we define a control vector
% \begin{equation}
% z_i=
% \left(
% z_i^{\mathrm{task}},
% z_i^{\mathrm{diff}},
% z_i^{\mathrm{theme}},
% z_i^{\mathrm{tables}}
% \right).
% \end{equation}
% $z_i^{\mathrm{task}}$ and $z_i^{\mathrm{tables}}$ jointly control SQL structure. The former samples from 14 task categories, including retrieval, filtering, aggregation, ranking, comparison, and trend analysis, and constrains the corresponding SQL operators; the latter determines whether the query uses one table or a multi-table combination. $z_i^{\mathrm{theme}}$ controls business themes such as players, teams, schedules, standings, and injuries. $z_i^{\mathrm{diff}}$ defines four difficulty levels, L1--L4, ranging from direct factual retrieval to multi-table joins, compound constraints, and multi-stage computation. Together, the controls shape SQL structure, domain semantics, and task difficulty.

For sample $i$, we define the control vector
\begin{equation}
z_i =
\left(
z_i^{\mathrm{task}},
z_i^{\mathrm{diff}},
z_i^{\mathrm{theme}},
z_i^{\mathrm{tables}}
\right).
\end{equation}
The components $z_i^{\mathrm{task}}$ and $z_i^{\mathrm{tables}}$ jointly constrain SQL structure. The former specifies one of 14 task categories, including retrieval, filtering, aggregation, ranking, comparison, and trend analysis, together with the corresponding SQL operators; the latter specifies whether the query involves a single table or multiple tables. The component $z_i^{\mathrm{theme}}$ selects a domain theme, such as players, teams, schedules, standings, or injuries. The component $z_i^{\mathrm{diff}}$ specifies one of four difficulty levels, L1-L4, ranging from direct factual retrieval to multi-table joins, compound constraints, and multi-stage computation. Together, these controls guide query synthesis across structural, semantic, and difficulty dimensions.

During initial data construction, $\mathcal{F}_0=\varnothing$, and synthesis is guided solely by the predefined controls. In subsequent rounds, the Feedback Engine analyzes execution failures and produces diagnostic findings $\mathcal{F}_t$ (Section~\ref{sec:eval}). These findings, including those concerning errors in join conditions, aggregation granularity, temporal filtering, and window ordering are translated into targeted constraints that guide SQL generation in round $t$:

\begin{equation}
\xi_i\sim p_{\theta_s}
\left(
\xi\mid\Sigma_i,z_i,\mathcal{F}_t
\right),
\end{equation}
where $\xi_i$ denotes a candidate SQL statement, $p_{\theta_s}$ the SQL-generation distribution, and $\Sigma_i$ the corresponding database schema. The scaling pipeline translates diagnostic findings into generation constraints, while the Feedback Engine provides the findings rather than generating training examples directly.

Each candidate SQL statement is checked for syntactic validity, schema consistency, and valid multi-table references, then deduplicated based on its normalized form. Retained statements are converted into natural-language questions:
\begin{equation}
q_i \sim p_{\theta_q}(q \mid \xi_i, \Sigma_i).
\end{equation}
The conversion must preserve all filtering, aggregation, ordering, and cardinality constraints without exposing table names, field names, or SQL terminology. The resulting questions are expressed naturally while remaining semantically verifiable against the underlying SQL.

\subsubsection{Graph-Guided Scaling}
Predefined task categories and difficulty levels do not fully capture the diversity of failures observed in production. Graph-Guided Scaling organizes diagnostic findings in $\mathcal{F}_t$ into a global error-attribution graph and samples local subgraphs that preserve relevant capability dependencies. These subgraphs guide question synthesis by linking each target root cause to its semantic, schema, and execution context.

\paragraph{Error Attribution Graph Construction.}
The global graph organizes individual failures into a reusable network of capability dependencies. For each failure, the Feedback Engine jointly examines the input query, execution trajectory, SQL results, judge decision, and relevant schema. The resulting diagnostic findings are aggregated into a heterogeneous graph:
\begin{equation}
\mathcal{G}_t =
\operatorname{Aggregate}\left(\mathcal{F}_t\right).
\end{equation}
The graph contains five node types: Root Cause, Sports Concept, Semantic Intent, Schema, and Error Pattern. Edges encode supporting-case links, semantic relations, and schema dependencies. Similar root causes are merged, while supporting cases, attribution confidence, and schema associations are retained and updated across rounds. This structure consolidates scattered failures into a reusable representation of recurring capability gaps.

\paragraph{Subgraph Sampling.}
Sampling operates on complete local capability-dependency structures rather than arbitrary node sets. Root causes are prioritized by supporting-case counts. For each selected root cause $r$, the system constructs candidate subgraphs that retain the associated concepts, semantic intents, schemas, error patterns, and dependencies, then samples subgraphs for the current synthesis round:
\begin{equation}
\mathcal{G}_{r,k}^{\mathrm{sub}} \sim
p\left(g \mid r, \mathcal{G}_t\right).
\end{equation}
The $k$-th candidate centered on root cause $r$ is represented by
\begin{equation}
\mathcal{G}_{r,k}^{\mathrm{sub}} =
\left\{
r,\,
c_k^{\mathrm{concept}},\,
i_k^{\mathrm{intent}},\,
\Sigma_k,\,
e_k^{\mathrm{pattern}}
\right\}.
\end{equation}
Candidate nodes are ranked by direct graph connections, shared failure cases, and shared tables. While keeping the target root cause fixed, the sampler varies domain concepts, semantic intents, and table combinations, producing structurally and semantically diverse subgraphs that target the same capability gap.

\paragraph{Subgraph-Guided Question Synthesis.}
Each sampled subgraph specifies structured constraints for question synthesis, including the target root cause, sports concept, semantic intent, relevant schema, and error pattern. Guided by these constraints, the generator varies entities, query conditions, difficulty levels, and linguistic forms to target a specific capability gap while maintaining semantic and structural diversity.

To reduce memorization and the risk of evaluation leakage, synthesis avoids reusing the original failed case, SQL statement, or entity combination. Instead, it recombines abstract diagnostic information across scenarios. Candidate questions are checked for answerability under the relevant schema, field consistency, constraint completeness, near duplication, and test-set overlap. A teacher model then generates complete execution trajectories for distillation, and a shared quality gate filters the resulting samples to form the final data increment.

\subsubsection{Agentic Trajectory Distillation and Training}

The two scaling strategies described above determine which questions should be included in the current evolution round. We next apply a shared trajectory-distillation and training pipeline to convert these questions into Agentic SQL supervision. For each synthesized question $q_i$, Qwen3.8-Max ~\citep{qwen3.8} serves as the teacher model and interacts with the  structured database environment. The teacher explores the schema, constructs and executes SQL, repairs execution errors, and generates an answer grounded in the returned results. The resulting training sample is

\begin{equation}
d_i =
\left(q_i,\tau_i,y_i\right),
\end{equation}

where $\tau_i$ records the complete sequence of tool calls, database observations, and intermediate corrections, and $y_i$ is the final evidence-grounded answer. Distilling the execution trajectory, rather than only the final answer, provides supervision for schema exploration, query planning, tool use, and error recovery.

A shared quality gate filters the distilled trajectories at both the outcome and execution levels. A retained sample must contain a complete question-tool call--tool result--answer chain, conform to the tool protocol, execute valid SQL, and support its final answer with database evidence. We discard trajectories containing syntax or field errors, timeouts, missing tool results, repeated SQL execution, abnormal interaction lengths, or incomplete answers. Questions are also normalized and deduplicated, and samples overlapping with the evaluation set are removed. The retained trajectories constitute the data increment $\Delta\mathcal{D}_t$ defined above.

SFT then incrementally updates $M_t$ using the mixed training set $\mathcal{D}_{t+1}$, following the model-update process formulated above~\citep{chen2023fireact,zeng2024agenttuning}. The loss is applied only to assistant-generated tokens:

\begin{equation}
\mathcal{L}_{\mathrm{SFT}}
=
-\sum_{d\in\mathcal{D}_{t+1}}
\sum_{j\in\mathcal{A}(d)}
\log p_{M_t}
\left(x_j\mid x_{<j}\right),
\end{equation}

where $\mathcal{A}(d)$ denotes assistant-token positions, including tool calls and final responses. Tool observations are excluded from the loss because they provide execution context rather than prediction targets~\citep{song2024agentbank}. This objective teaches the model to preserve the interaction protocol while improving the query and recovery behaviors targeted in the current evolution round.

\subsection{Post-RL}
\label{subsec:model-loop-post-rl}
Post-RL is performed after the iterative model-evolution loop. Starting from $M_T$, RL directly optimizes final Agentic SQL task outcomes~\citep{ma2026sql,feng2026retool} and produces $M_{\mathrm{final}}$. We use GSPO~\citep{gspo}: a group of trajectories is sampled for each prompt, rewards are normalized within the group to obtain advantages, and a clipped objective is optimized using sequence-level importance ratios. The complete objective and sampling configuration are provided in Appendix~\ref{app:agentic-sql-rl}; this section focuses on reward design.

\paragraph{Reward Design.}
Agentic SQL tasks vary in their degree of verifiability. Some admit a unique executable reference, whereas others must be evaluated by whether the execution trajectory fulfills the underlying query intent. No single evaluation oracle is suitable for both settings: an LLM-based judge introduces unnecessary uncertainty for directly verifiable tasks, while execution-based verification alone cannot adequately assess valid open-ended analyses. We therefore choose the strongest available oracle for each task type and align the reward structure with the task structure. Rewards remain sparse terminal scalars at the trajectory level; training only forwards and normalizes them within each group, without a learned reward model or step-level shaping. Table~\ref{tab:rl-task-types} summarizes the design.

\begin{table}[htbp]
  \centering
  \caption{Reward design for the two Agentic SQL task types.}
  \label{tab:rl-task-types}
  \small
  \linespread{1.0}\selectfont          % 单元格内部折行行距
  \renewcommand{\arraystretch}{1.35}    % 行与行之间的留白
  \begin{tabular}{>{\raggedright\arraybackslash}p{0.18\linewidth}
                  >{\raggedright\arraybackslash}p{0.28\linewidth}
                  >{\raggedright\arraybackslash}p{0.28\linewidth}
                  >{\raggedright\arraybackslash}p{0.12\linewidth}}
    \toprule
    Task Type & Verification Oracle & Reward Form & Granularity \\
    \midrule
    Execution-verifiable & Database executor: result equivalence & Stage-weighted execution equivalence & Stage \\
    Open-ended analysis & LLM judge: three-dimensional assessment & Fraction of successful repeated judgments & Trajectory \\
    \bottomrule
  \end{tabular}
\end{table}

\paragraph{Execution-Verifiable Tasks.}
Correctness can be determined objectively by the database executor, so the reward is anchored in result equivalence between predicted and reference queries. To reflect composite tasks, the reward is defined over stages:
\begin{equation}
R_{\mathrm{exec}} = \sum_{s=1}^{C_{\mathrm{exec}}} w_s \cdot \mathbb{1}\left[\mathrm{exec}(\hat{y}_s) \equiv \mathrm{exec}(y_s^*)\right],
\qquad
w_s \geq 0,\ \textstyle\sum_{s=1}^{C} w_s = 1.
\end{equation}
$\hat{y}_s$ and $y_s^*$ are the predicted and reference queries for stage $s$, $w_s$ is the stage weight, $C$ is the number of stages defined by the task, and $C_{\mathrm{exec}}$ is the number of executable stages. A single-stage task reduces to a binary execution-equivalence test. Multi-stage rewards accumulate in order with weights reflecting stage importance, distinguishing partial progress while reserving the maximum reward for complete solutions. Appendix~\ref{app:agentic-sql-rl} gives the instantiated weights and normalization rules.

\paragraph{Open-Ended Analysis Tasks.}
These tasks have no unique reference SQL. Correctness is a trajectory-level property: a single execution result cannot determine whether all constraints were covered, the queries were valid, or the exploration converged. An LLM judge therefore evaluates intent coverage, query validity, and trajectory efficiency. To reduce single-call variance, the reward is the success fraction over repeated independent judgments:
\begin{equation}
R_{\mathrm{judge}} = \frac{1}{k} \sum_{i=1}^{k} \mathbb{1}[\text{resolved}_i].
\end{equation}
The judgment is based on semantic equivalence and does not penalize immaterial boundary-expression differences such as $>$ versus $\geq$. Missing required constraints, adding or removing filters that alter the result, substituting an approximate metric, or answering only part of the question causes a failure. Appendix~\ref{app:agentic-sql-rl} specifies the number of judgments and judge configuration.

\paragraph{Format Gate.}
Outcome rewards alone may assign a positive reward to a trajectory whose interaction format has degraded. Before training, the raw reward is multiplied by a binary gate, $R_{\mathrm{final}}=R_{\mathrm{raw}}\cdot G_{\mathrm{format}}$. The gate checks whether the text preceding the first tool call in each round follows the required format. It is a necessary condition rather than a bonus and is applied only when $R_{\mathrm{raw}}>0$, so it can reject a malformed positive trajectory but cannot turn a zero reward into a positive one.

\subsection{Experiments}
\label{subsec:model-loop-experiments}

\subsubsection{Experimental Setup}

\begin{table}[htbp]
    \centering
    \caption{Performance comparison on Sports Agentic SQL task. Best and second-best results are shown in \textbf{bold} and \underline{underlined}, respectively.}
    \label{tab:nba-model-performance}
    \begin{tabular}{lcc}
      \toprule
      Model & Sports-Normal & Sports-Hard \\
      \midrule
      Qwen3.8-Max~\citep{qwen3.8} & 94.5\% & 72.3\% \\
      GLM-5.2~\citep{zeng2026glm} & 92.5\% & 60.0\% \\
      DeepSeek-V4-Pro~\citep{xu2026deepseek} & 89.6\% & \textbf{76.0\%} \\
      Qwen3.6-35B-A3B~\citep{qwen3.6-35b-a3b} & 83.5\% & 45.7\% \\
      Qwen3.6-35B-A3B-Evolve & \underline{96.7\%} & 70.7\% \\
      Qwen3.6-35B-A3B-Evolve + RL & \textbf{97.8\%} & \underline{74.0\%} \\
      \bottomrule
    \end{tabular}
  \end{table}

We evaluate the model on two task families. \textbf{Sports Agentic SQL} assesses domain-specific capability over sports databases used in the evaluated application, with Sports-Normal and Sports-Hard subsets covering routine queries and more complex sports analytical tasks. \textbf{General Agentic SQL} assesses generalization to external databases and domains using LiveSQLBench~\citep{livesqlbench2025} and BIRD-Interact~\citep{huo2026bird}. Appendix~\ref{app:agentic-sql-setup} describes the datasets and evaluation settings.

To reflect production latency constraints, we disable explicit thinking mode for all evaluated models and retain this configuration on the general benchmarks. For Sports Agentic SQL, we report task correctness, end-to-end latency, and average tool-call rounds. Each test sample is evaluated three times, and the reported results are averaged across the three runs. For General Agentic SQL, we report task reward and average tool-call rounds.

\subsubsection{Model Performance Evaluation}

As shown in Table~\ref{tab:nba-model-performance}, Qwen3.6-35B-A3B-Evolve outperforms the base model by 13.2\%  on Sports-Normal and 25.0\% on Sports-Hard, with a larger gain on the more challenging subset. Post-RL yields additional gains of 1.1 and 3.3 percentage points, respectively.

\begin{table}[htbp]
  \centering
  \caption{Performance on General Agentic SQL. Best and second-best results are shown in \textbf{bold} and \underline{underlined}, respectively.}
  \label{tab:agentic-sql-benchmark-performance}
  \small
  \setlength{\tabcolsep}{4pt}
  \resizebox{\linewidth}{!}{%
  \begin{tabular}{lccccc}
    \toprule
    & \multicolumn{3}{c}{LiveSQLBench} & \multicolumn{2}{c}{BIRD-Interact} \\
    \cmidrule(lr){2-4}\cmidrule(lr){5-6}
    Model & Reward (Lite) & Reward (Full) & Reward (Large) & Reward (Lite) & Reward (Full) \\
    \midrule
    Qwen3.8-Max & 46.7\% & \underline{34.3\%} & \underline{32.3\%} & 23.0\% & 21.9\% \\
    Kimi-K3~\citep{team2026kimi} & \textbf{47.8\%} & 33.8\% & \textbf{32.7\%} & 26.8\% & 21.9\% \\
    GLM-5.2 & 44.8\% & 30.3\% & 26.3\% & 19.7\% & 15.7\% \\
    Qwen3.6-35B-A3B & 41.5\% & 29.3\% & 24.8\% & 15.2\% & 13.9\% \\
    Qwen3.6-35B-A3B-Evolve & 46.3\% & 32.8\% & \underline{32.3\%} & \underline{28.4\%} & \underline{23.4\%} \\
    Qwen3.6-35B-A3B-Evolve + RL & \underline{47.4\%} & \textbf{37.2\%} & 31.9\% & \textbf{54.0\%} & \textbf{41.0\%} \\
    \bottomrule
  \end{tabular}%
  }
\end{table}

As shown in Table~\ref{tab:agentic-sql-benchmark-performance}, Qwen3.6-35B-A3B-Evolve outperforms the base model across all five General Agentic SQL metrics, with gains of 4.8, 3.5, and 7.5 percentage points on LiveSQLBench Lite, Full, and Large, respectively, and 13.2 and 9.5 percentage points on BIRD-Interact Lite and Full. Post-RL further improves four of the five metrics, with particularly large gains on BIRD-Interact, although performance on LiveSQLBench Large declines slightly.

\subsubsection{Model Efficiency Evaluation}
\begin{figure*}[htbp]
    \centering
    \begin{minipage}[t]{0.485\textwidth}
      \vspace{0pt}
      \centering
      \includegraphics[height=0.68\linewidth,keepaspectratio]{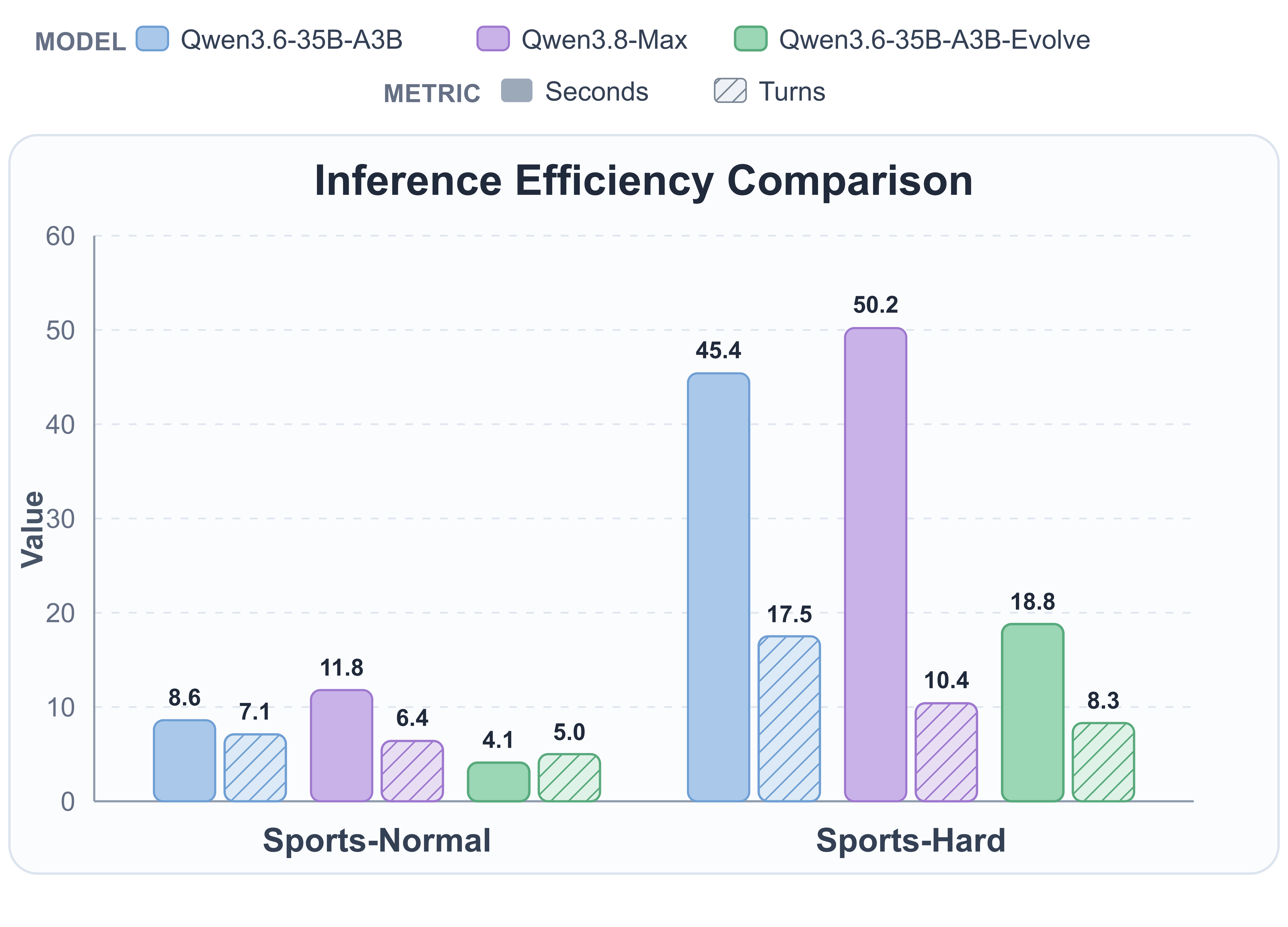}
      \captionof{figure}{Efficiency on Sports Agentic SQL in terms of latency and interaction turns.}
      \label{fig:agentic-sql-efficiency}
    \end{minipage}
    \hfill
    \begin{minipage}[t]{0.46\textwidth}
      \vspace{0pt}
      \centering
      \includegraphics[height=0.68\linewidth,keepaspectratio]{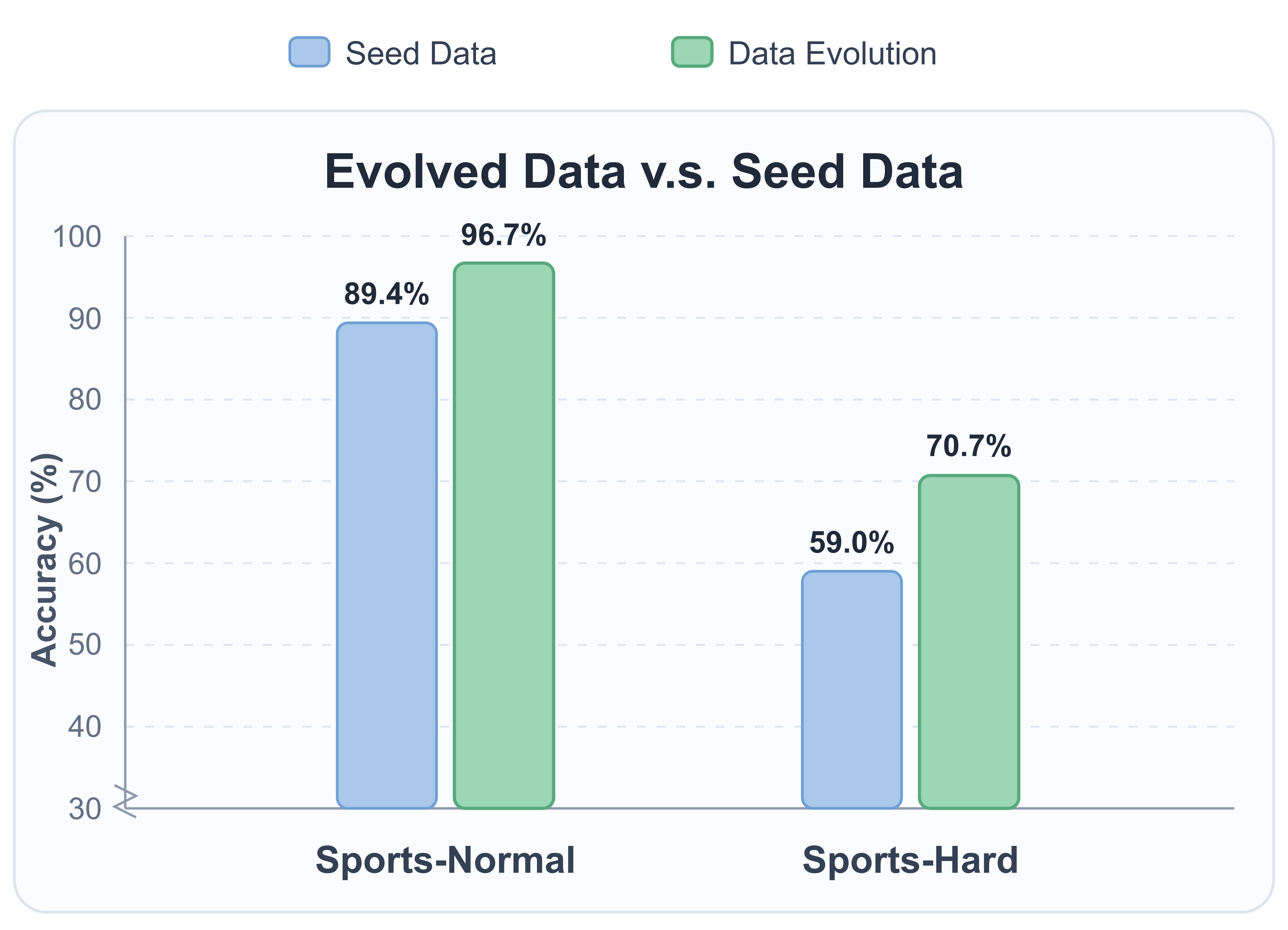}
      \captionof{figure}{Ablation of data-evolution strategies on Sports-Normal and Sports-Hard.}
      \label{fig:question-synthesis-ablation}
    \end{minipage}
  \end{figure*}

Figure~\ref{fig:agentic-sql-efficiency} compares end-to-end latency and average tool-call rounds on Sports Agentic SQL. Relative to Qwen3.6-35B-A3B, the evolved model reduces latency by 52.3\% on Sports-Normal and 58.6\% on Sports-Hard, and tool-call rounds by 29.6\% and 52.6\%, respectively. These results show that accuracy gains are accompanied by lower latency and fewer tool interactions. Table~\ref{tab:general-tool-call-rounds} reports tool-use efficiency on General Agentic SQL.

\subsubsection{Ablation Study}

\paragraph{Effect of Data Evolution.}
To demonstrate the effectiveness of data evolution, we hold the base model, training configuration, and evaluation protocol fixed, and compare models trained on the curated seed dataset and the evolved dataset. As shown in Figure~\ref{fig:question-synthesis-ablation}, Data Evolution improves accuracy from 89.4\% to 96.7\% on Sports-Normal and from 59.0\% to 70.7\% on Sports-Hard, yielding gains of 7.3 and 11.7 percentage points, respectively. The larger improvement on Sports-Hard is consistent with the evolved dataset’s broader coverage of multi-table joins, compound constraints, and long-tail SQL structures that are less prevalent in the curated seed dataset. These results suggest that data evolution improves performance on both routine queries and more complex Agentic SQL tasks.

\begin{figure}[htbp]
  \centering
  \includegraphics[width=0.9\linewidth]{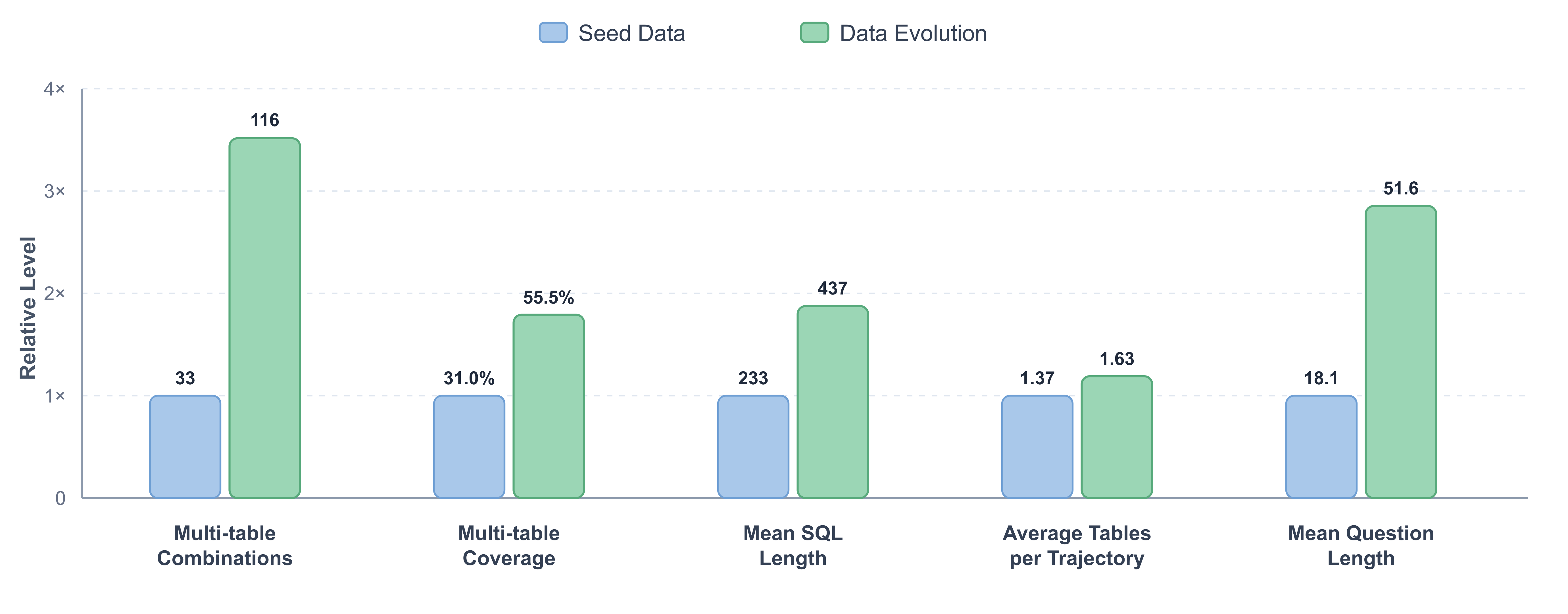}
  \caption{Multidimensional comparison of the curated seed dataset and the evolved dataset. Bar heights are normalized to the seed dataset, with original values reported above each bar.}
  \label{fig:training-data-comparison}
\end{figure}

\paragraph{Statistics of the Evolved Data.}
Figure~\ref{fig:training-data-comparison} compares the curated seed dataset with the evolved dataset across several dimensions. The evolved dataset contains approximately 3.5 times as many multi-table combinations, while its multi-table coverage and mean SQL length are nearly twice as high and its mean question length is approximately three times greater. These results show that data evolution expands the coverage of cross-table relationships, compound constraints, and compositional queries beyond the patterns represented in the seed dataset.

\begin{figure}[htbp]
  \centering
  \includegraphics[width=0.9\linewidth]{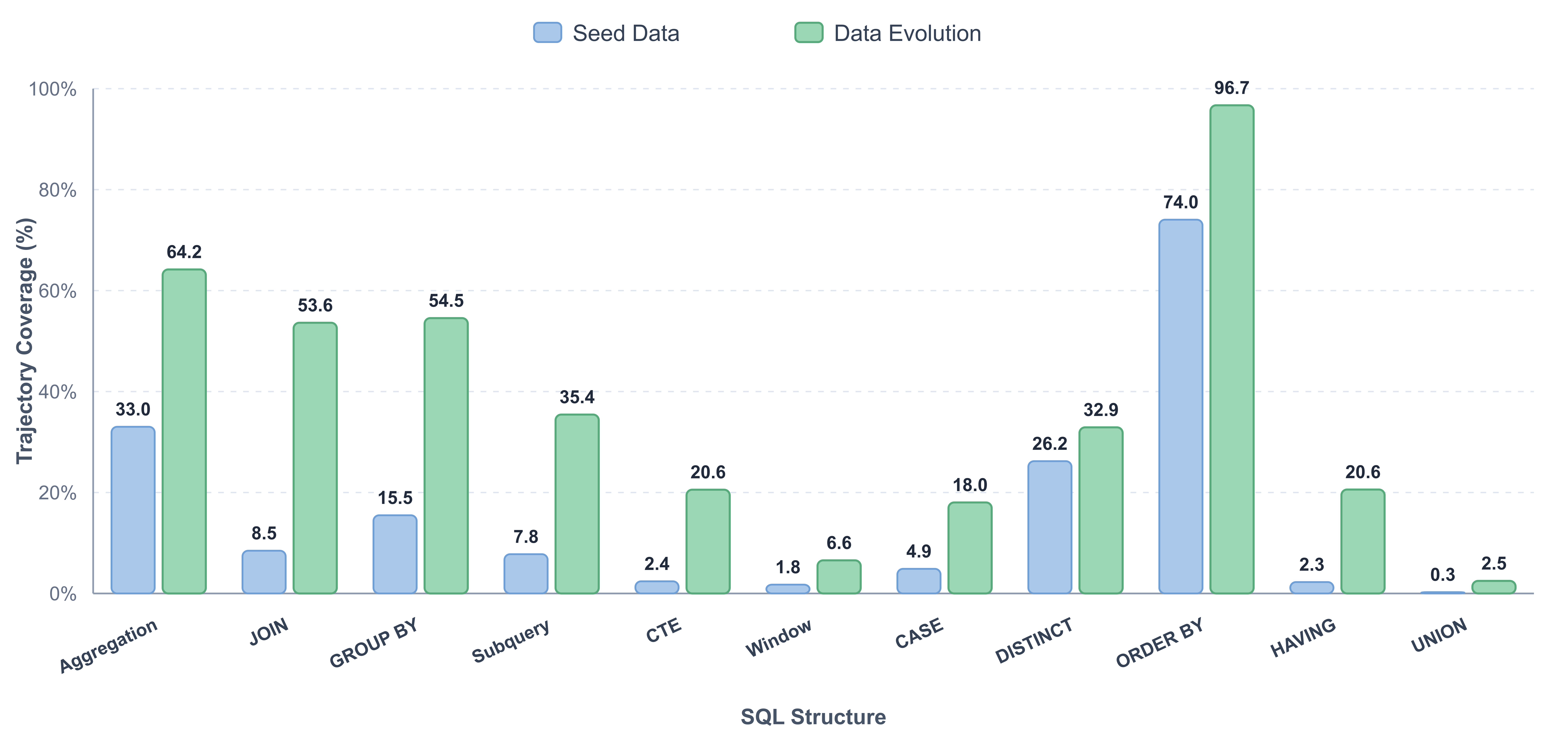}
  \caption{SQL-structure coverage of the curated seed dataset and the evolved dataset. The evolved dataset provides broader coverage of both common operators and long-tail complex structures.}
  \label{fig:sql-coverage-comparison}
\end{figure}

Figure~\ref{fig:sql-coverage-comparison} further examines SQL-structure coverage. The evolved dataset substantially increases the prevalence of joins, subqueries, common table expressions, window functions, and other complex operators. Several long-tail structures occur four to nine times as frequently as in the seed dataset, while common aggregation and ordering operations also receive broader coverage. Data evolution therefore reduces the initial dataset’s bias toward simple and frequent tasks and adapts the training distribution to evolving schemas and diagnosed capability gaps. Together with the observed improvements in accuracy, generalization, and execution efficiency, these findings demonstrate that failure diagnosis can be translated into targeted training data, enabling the model to continually refine its training distribution rather than relying on a one-time data-construction process.

%% file: Sports_Agent_Anonymization/Harness-loop.tex
\section{Harness Self-Evolution Loop}
\label{sec:harness-loop}

Improving Sports Agent requires the harness to adapt as tasks, data, and model capabilities evolve. As discussed in Section~\ref{sec:introduction}, these changes can disrupt skill activation, invalidate tool contracts and domain guidance, or make existing components redundant as models internalize their functionality. Addressing these failures requires targeted edits to policies, content, activation conditions, and component structure. Reliable harness evolution therefore links runtime evidence, failure attribution, bounded edits, and version validation in an iterative loop, ensuring that persistent changes are supported by evidence and verified before release.

We develop a harness-evolution framework in which the Feedback Engine serves as the diagnostic entry point, typed repair intents as the unit of evolution, and independent auditing and execution replay as the validation mechanisms. The Feedback Engine assesses relevant execution evidence and diagnoses harness-side capability gaps (Section~\ref{sec:eval}), translating them into structured repair intents. Related intents accumulate across evolution events and trigger bounded edits only after receiving sufficient support. Each candidate edit undergoes an independent audit and execution verification, after which validated edits are jointly replayed for final selection. Finally, evidence-based consolidation and a non-inferiority guard control configuration growth and produce the next harness version.

Section~\ref{sec:harness-overview} presents the overall process and unit of evolution. Section~\ref{sec:harness-intent} describes persistent intent aggregation and bounded edit synthesis. Section~\ref{sec:harness-verification} introduces independent auditing, execution verification, and final selection. Section~\ref{sec:harness-consolidation} explains evidence-based consolidation and non-inferiority gating.

\subsection{Turning Accumulated Evidence into Coordinated Repairs}
\label{sec:harness-overview}

The Sports Agent harness determines how the agent plans tasks, invokes tools, organizes context, and produces responses~\citep{guo2026question}. It consists of behavioral policies, solution procedures, tool contracts, reusable capability content (skills), and the activation relationships among them. This section instantiates the harness update operator $\mathcal{U}_H$ from Section~\ref{sec:overview} as an offline evolution loop driven by Feedback Engine outputs.

\begin{figure}[htbp]
  \centering
  \includegraphics[width=\linewidth]{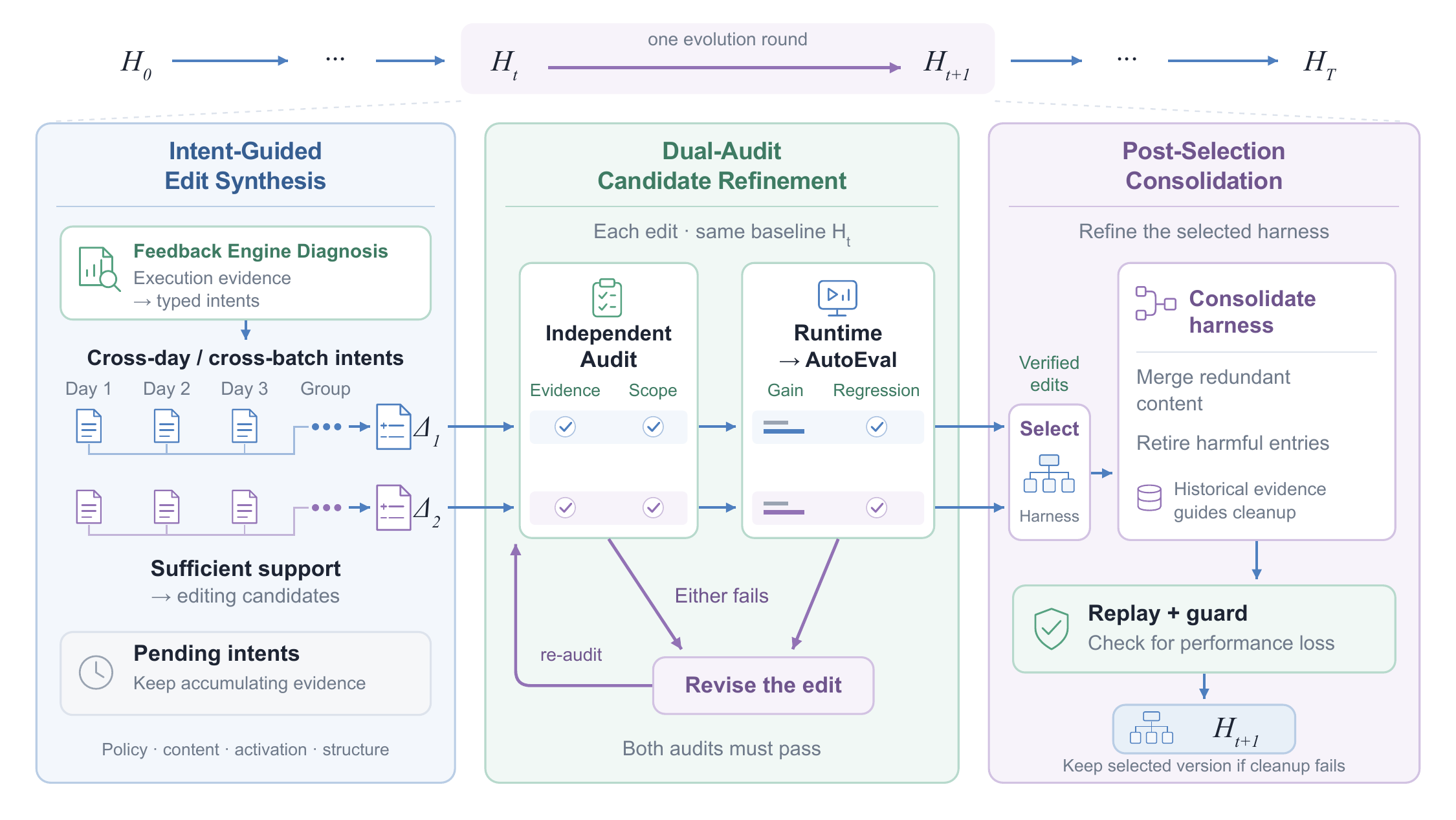}
  \caption{Harness self-evolution. Repair intents accumulate across events and produce bounded edit candidates after reaching a support threshold. Each candidate undergoes independent audit and runtime verification. Validated edits are jointly replayed to produce $H_t^{\mathrm{win}}$, followed by evidence-based consolidation and a non-inferiority guard for $H_{t+1}$.}
  \label{fig:harness-loop}
\end{figure}

Three difficulties separate observed degradation from a reliable repair. First, evidence for a persistent change is often sparse~\citep{liu2026rethinking}; an individual failure may result from transient execution variance or a data anomaly, so directly editing the harness risks overfitting. Second, root causes are heterogeneous~\citep{cemri2025mast}. A missing global policy, incorrect capability content, failed capability activation, and an unclear component boundary require different targets and edit operations. Third, individually useful edits may conflict when combined, and an append-only update strategy steadily increases redundancy and complexity. Figure~\ref{fig:harness-loop} presents the resulting loop.

One evolution round is represented as
\begin{equation}
  \underbrace{(H_t,\Omega_t)\rightarrow\mathcal{I}_t\rightarrow\mathcal{C}_t^{\mathrm{ready}}\rightarrow\{\Delta_{t,j}\}}_{\text{(i) Intent-Guided Edit Synthesis}}
  \underbrace{\rightarrow\mathcal{V}_t\rightarrow H_t^{\mathrm{win}}}_{\text{(ii) Dual Audit}}
  \underbrace{\rightarrow H_t^{\mathrm{cons}}\rightarrow(H_{t+1},\Omega_{t+1})}_{\text{(iii) Post-Selection Consolidation}}.
  \label{eq:harness-round}
\end{equation}
$H_t$ is the current harness, $\Omega_t$ is cross-event evolution memory, and $\mathcal{I}_t$ contains repair intents emitted by the Feedback Engine---the concrete form of the harness evolution signal $g^H$ introduced in Section~\ref{sec:overview}. $\mathcal{C}_t^{\mathrm{ready}}$ contains intent groups that meet the evidence threshold; $\Delta_{t,j}$ is a local edit candidate; $\mathcal{V}_t$ is the validated edit set; $H_t^{\mathrm{win}}$ is the version selected after joint replay; and $H_t^{\mathrm{cons}}$ is the consolidated candidate. Repairs are decomposed by root-cause mechanism and target component rather than attached to individual failures. Each edit is audited and verified independently so that gains are attributable and failures remain isolated. Final selection and consolidation operate on the complete harness to control interactions among edits.

\paragraph{Evolution Granularity.}
The unit of evolution is an \emph{evolution event}; $t$ indexes events. Each event comprises relevant execution evidence from a predefined observation window, typically organized as a daily batch, together with the corresponding Feedback Engine assessments. An event can produce at most one update $H_t\rightarrow H_{t+1}$. A deterministic split partitions the event sample set into training and held-out subsets:

\begin{equation}
  \mathcal{Q}_t=\mathcal{Q}_t^{\mathrm{tr}}\cup\mathcal{Q}_t^{\mathrm{te}},\qquad
  \mathcal{Q}_t^{\mathrm{tr}}\cap\mathcal{Q}_t^{\mathrm{te}}=\varnothing .
\end{equation}
$\mathcal{Q}_t^{\mathrm{tr}}$ is used for attribution, intent generation, and edit drafting. $\mathcal{Q}_t^{\mathrm{te}}$ remains isolated from these steps and is used only for candidate gains, regression checks, final selection, and non-inferiority gating. An event may contain multiple epochs, each repeating attribution, editing, and validation from the previous epoch's output. Cross-event state---including pending intents, component usage and impact statistics, and retired entries---is stored in $\Omega_t$. This allows infrequent problems to accumulate evidence over multiple events.

\subsection{Intent Aggregation and Edit Synthesis}
\label{sec:harness-intent}

\subsubsection{Persistent Intent Aggregation}
\label{sec:harness-aggregation}

A typed repair intent describes the attributed layer of a failure, the proposed edit action, its target, supporting cases, and rationale.

\paragraph{Layer Attribution.}
To handle prompts and skills within a common framework, attribution identifies the harness layer responsible for a failure using the actual system prompt, activated capabilities, and execution trajectory. A \textbf{Prompt Gap} is a missing cross-task behavioral rule that is absent from the current system prompt and must apply globally. \textbf{Skill Content} denotes a missing or incomplete reusable capability, including errors in domain knowledge, retrieval procedures, or statistical conventions within an activated skill. \textbf{Skill Activation} indicates that the required capability exists but was not retrieved for the request. \textbf{Skill Harm} indicates that an activated capability actively misdirected the solution.

Cases in which an existing rule was not followed, no usable data exist after a reasonable execution, or an evaluator contradicts the evidence are classified as non-harness defects. They do not produce harness edits, although they remain in the evolution memory for analysis. Consistent with Section~\ref{sec:eval}, failures to follow an available rule are better addressed through model training or execution constraints. A case may carry both prompt- and skill-level attributions; the corresponding intents enter separate aggregation paths rather than being treated as mutually exclusive.

\paragraph{Aggregation.}
The system combines intents from different minibatches with pending intents retained in $\Omega_t$. Aggregation is stratified by attribution layer. Skill intents are grouped primarily by edit type and target component; creation intents, whose target does not yet exist, are grouped by question type. Prompt intents are clustered by the semantics of the proposed global rule and must be supported by multiple question types, preventing a specialized deficiency from being promoted into a global policy. Supporting cases, failed dimensions, proposals, and root-cause statistics are merged into single-layer intent groups $\mathcal{C}_j^{\mathrm{intent}}$.

Support is the number of distinct \texttt{event:case\_id} values in a group, denoted by $\operatorname{supp}(\mathcal{C}_j^{\mathrm{intent}})$. This allows evidence to accumulate across events while preventing duplicate cases from being counted multiple times. Groups meeting threshold $\lambda_{\mathrm{supp}}$ become ready:
\begin{equation}
  \mathcal{C}_t^{\mathrm{ready}}=
  \left\{\mathcal{C}_j^{\mathrm{intent}} \mid
  \operatorname{supp}(\mathcal{C}_j^{\mathrm{intent}})\geq \lambda_{\mathrm{supp}}\right\}.
\end{equation}
Groups with insufficient evidence remain pending; groups whose targets are obsolete or whose waiting time exceeds a limit are expired. The threshold establishes an explicit boundary between a one-off failure and a persistent system change.

Before synthesis, ready groups are prioritized by support, failure severity, and diagnostic confidence, subject to a per-round proposal budget. Preserve clauses extracted from successful cases are attached as behavioral constraints. This design gives infrequent issues time to accumulate evidence while limiting the scope of each update round.

\subsubsection{Bounded Edit Synthesis}
\label{sec:harness-synthesis}

Each ready intent group produces an edit candidate $\Delta_{t,j}$. Prior work has used LLM feedback to revise textual configurations~\citep{yuksekgonul2025optimizing,agrawal2026gepa}; our candidates are instead bound to an attribution layer, preserve constraints, and an edit budget. Each candidate consists of explicitly scoped actions and is associated with the baseline $H_t$, target intent, protection constraints, and allowed edit magnitude. Table~\ref{tab:harness-edit-space} defines the edit space.

\begin{table}[t]
  \centering
  \caption{Harness edit space.}
  \label{tab:harness-edit-space}
  \linespread{1.0}
  \fontsize{9}{10}\selectfont
  \renewcommand{\arraystretch}{1.5}
  \begin{tabularx}{\linewidth}{l l X X X}
    \toprule
    \textbf{Type} & \textbf{Layer} & \textbf{Attribution} & \textbf{Primary Target} & \textbf{Typical Operation} \\
    \midrule[0.9pt]
    Policy & Prompt & Prompt gap & Global behavior and tool contracts & Add, rewrite, or narrow a rule \\
    \arrayrulecolor{black!20}
    \midrule[0.3pt]
    \arrayrulecolor{black}
    Content & Skill & Missing or incomplete capability & Specialized knowledge, procedures, and examples & Create, extend, or correct content \\
    \arrayrulecolor{black!20}
    \midrule[0.3pt]
    \arrayrulecolor{black}
    Activation & Skill & Capability not activated & Descriptions, triggers, and task mappings & Revise retrieval and routing conditions \\
    \arrayrulecolor{black!20}
    \midrule[0.3pt]
    \arrayrulecolor{black}
    Structure & Skill & Harmful capability or cross-event overlap & Component boundaries, indices, and dependencies & Merge, split, retire, or rebuild an index \\
    \bottomrule
  \end{tabularx}
\end{table}

Each ready group maps to an edit type according to its attribution. Prompt gaps yield Policy candidates over the system prompt. The three skill-side attributions yield Content, Activation, or Structure candidates. For example, incomplete rules within a correctly retrieved skill require a Content edit, whereas an existing skill that was not retrieved requires an Activation edit. Structure edits may also be initiated during consolidation from cross-event evidence of overlap or persistent harm (Section~\ref{sec:harness-consolidation}).

Different attribution layers impose different constraints. A Policy edit must express a durable, cross-task behavioral principle and cannot introduce entity-specific values, numerical answers, or task templates. Content and Activation edits are scoped to one task family and cannot move general behavior into a specialized skill. These boundaries prevent duplicate repairs across prompts and skills and protect global rules from specialized knowledge. Candidate isolation also reduces experimental confounding.

An edit candidate is only a proposal. During verification, $\Delta_{t,j}$ is applied to the fixed baseline $H_t$ in an isolated environment, producing $\widetilde{H}_{t,j}$. Sharing the same baseline makes gains from different edits directly comparable.

\subsection{Dual-Audit Candidate Refinement}
\label{sec:harness-verification}

Each edit candidate undergoes two checks: an independent static audit followed by dynamic execution verification.

\paragraph{Independent Audit.}
Following LLM-as-a-Judge practice~\citep{zheng2023llmjudge,gu2026survey,liu2023geval,kim2024prometheus}, an independent model checks whether the proposed edit is supported by evidence, remains within its attributed scope, and preserves structural, interface, dependency, and protection constraints. The audit is designed to reject hard-coded examples, unsupported rule expansion, and internal inconsistency.

\paragraph{Execution Verification.}
After an edit passes audit, the runtime executes both $H_t$ and $\widetilde{H}_{t,j}$. The Feedback Engine compares paired gains on target cases, regressions on non-target tasks, protected behavior, and actual capability activation. Only candidates passing both checks enter
\begin{equation}
  \mathcal{V}_t=
  \left\{\Delta_{t,j}\ \middle|\
    \operatorname{Audit}(\Delta_{t,j})=\mathrm{pass}
    \ \land\
    \operatorname{Verify}(H_t,\widetilde{H}_{t,j})=\mathrm{pass}
  \right\}.
\end{equation}

\paragraph{Bounded Revision.}
If either check fails, the edit may be revised using the returned feedback and resubmitted under a new revision identifier. Revision terminates after a fixed number of attempts or when the execution budget is exhausted. Failure reasons and unresolved intents are retained in evolution memory for future events.

\paragraph{Final Selection.}
Validated edits are jointly applied to $H_t$ to form a candidate version, which is compared with the baseline through full replay on $\mathcal{Q}_t^{\mathrm{te}}$. $\operatorname{score}(H)$ is the mean Feedback Engine score on the held-out set. Because metric differences must be interpreted relative to evaluator noise~\citep{miller2024adding}, a candidate is accepted as $H_t^{\mathrm{win}}$ only when its gain exceeds the noise threshold. Differences within the threshold are treated as ties, in which case the round falls back to $H_t$. Final selection confirms that locally validated improvements remain beneficial when combined in the complete harness.

\subsection{Post-Selection Consolidation}
\label{sec:harness-consolidation}

\subsubsection{Evidence-Based Consolidation}
\label{sec:harness-consolidation-evidence}

After editing and final selection, the system controls the duplication and accumulation introduced by long-term evolution. Consolidation searches the selected version for highly overlapping entries, components with similar function, and entries with persistent evidence of harm. Components that have not been used for an extended period are marked dormant rather than immediately removed, protecting rare but valid capabilities.

Consolidation candidates are identified from $\Omega_t$. A merge requires both content similarity and sufficient related-failure evidence; retirement requires sustained evidence of harm. If no candidate meets these criteria, consolidation terminates without additional generation or evaluation.

For eligible candidates, a consolidation plan is generated and applied to a copy of $H_t^{\mathrm{win}}$. Guardrails limit edit magnitude and check that existing constraints and activation information are preserved, producing $H_t^{\mathrm{cons}}$. Following ACE~\citep{zhang2026agentic}, consolidation does not search for a new optimum; it removes clear redundancy and persistent harm while preserving behavior as closely as possible.

\subsubsection{Non-Inferiority Guard}
\label{sec:harness-noninferiority}

Because multiple consolidation operations are applied together, their combined effect cannot be inferred from the evidence for each individual operation. Merging or retiring components may also damage infrequent but useful behavior. The system therefore replays $H_t^{\mathrm{cons}}$ on the same held-out set used for final selection and compares it with $H_t^{\mathrm{win}}$:
\begin{equation}
  \operatorname{score}\bigl(H_t^{\mathrm{cons}}\bigr)\geq
  \operatorname{score}\bigl(H_t^{\mathrm{win}}\bigr)-\epsilon_{\mathrm{NI}},
  \label{eq:harness-noninferiority}
\end{equation}
where $\epsilon_{\mathrm{NI}}\geq 0$ is the maximum acceptable performance loss. By default, it equals the evaluation-noise threshold used in final selection. Comparing against $H_t^{\mathrm{win}}$, rather than the original $H_t$, separates the effect of editing from the effect of consolidation.

Final selection and consolidation use the same noise scale in opposite directions. New or modified behavior must improve by more than the noise threshold; consolidation primarily reduces redundancy and is accepted when it is not meaningfully worse. Requiring a significant score gain from consolidation would reject nearly every cleanup and force the harness to grow monotonically. The non-inferiority criterion instead treats a smaller configuration as an improvement when its performance is statistically indistinguishable, while $\epsilon_{\mathrm{NI}}$ bounds the price paid for simplification.

The next version is
\begin{equation}
  H_{t+1}=
  \begin{cases}
    H_t^{\mathrm{cons}}, & \text{if the non-inferiority guard passes},\\
    H_t^{\mathrm{win}},  & \text{otherwise}.
  \end{cases}
\end{equation}
Pending intents, failure reasons, and component usage and impact statistics are written back to produce $\Omega_{t+1}$.

\begin{algorithm}[t]
  \caption{One round of harness self-evolution}
  \label{alg:harness-round}
  \begin{algorithmic}[1]
    \Require Current harness $H_t$, memory $\Omega_t$, event split $(\mathcal{Q}_t^{\mathrm{tr}},\mathcal{Q}_t^{\mathrm{te}})$, and repair intents $\mathcal{I}_t$ derived from $\mathcal{Q}_t^{\mathrm{tr}}$
    \Ensure Next harness $H_{t+1}$ and updated memory $\Omega_{t+1}$
    \State Aggregate $\mathcal{I}_t$ with pending intents in $\Omega_t$ by attribution layer and target (or by task type for creation intents)
    \State $\mathcal{C}_t^{\mathrm{ready}} \gets \{\mathcal{C}_j^{\mathrm{intent}} \mid \operatorname{supp}(\mathcal{C}_j^{\mathrm{intent}})\geq\lambda_{\mathrm{supp}}\}$; retain other groups as pending or expired
    \State Rank ready groups by support, severity, and confidence; truncate to the proposal budget and attach preserve clauses
    \State $\mathcal{V}_t \gets \emptyset$
    \ForAll{$\mathcal{C}_j^{\mathrm{intent}} \in \mathcal{C}_t^{\mathrm{ready}}$}
      \State Select an edit type from the attribution and synthesize bounded edit $\Delta_{t,j}$
      \Repeat
        \State Apply $\Delta_{t,j}$ to isolated baseline $H_t$ to obtain $\widetilde{H}_{t,j}$
        \If{$\operatorname{Audit}(\Delta_{t,j})=\mathrm{pass}$ \textbf{and} $\operatorname{Verify}(H_t,\widetilde{H}_{t,j})=\mathrm{pass}$}
          \State $\mathcal{V}_t \gets \mathcal{V}_t \cup \{\Delta_{t,j}\}$; \textbf{break}
        \EndIf
        \State Revise $\Delta_{t,j}$ from audit or verification feedback
      \Until{the revision or execution budget is exhausted}
    \EndFor
    \State Jointly replay a version constructed from $\mathcal{V}_t$ and select $H_t^{\mathrm{win}}$; set $H_t^{\mathrm{win}}=H_t$ if no gain is established
    \If{$\Omega_t$ contains entries meeting merge or retirement criteria}
      \State Apply consolidation to a copy of $H_t^{\mathrm{win}}$ to obtain $H_t^{\mathrm{cons}}$
      \If{$\operatorname{score}(H_t^{\mathrm{cons}})\geq\operatorname{score}(H_t^{\mathrm{win}})-\epsilon_{\mathrm{NI}}$}
        \State $H_{t+1}\gets H_t^{\mathrm{cons}}$
      \Else
        \State $H_{t+1}\gets H_t^{\mathrm{win}}$
      \EndIf
    \Else
      \State $H_{t+1}\gets H_t^{\mathrm{win}}$
    \EndIf
    \State Write pending intents, failure reasons, and usage statistics to $\Omega_{t+1}$
  \end{algorithmic}
\end{algorithm}

Algorithm~\ref{alg:harness-round} summarizes one complete round. In summary, the Harness Self-Evolution Loop uses accumulated evidence to decide whether a change is warranted, root-cause attribution to determine where the change belongs, independent validation and full replay to decide whether it should be retained, and consolidation under a non-inferiority guard to control long-term complexity. Diagnostic signals from the Feedback Engine are thereby converted into coordinated and verifiable harness updates.

\subsection{Case Study: From Repair Intents to Layer-Specific Updates}
\label{sec:harness-case-study}

\begin{figure}[!htb]
  \centering
  \includegraphics[width=\linewidth]{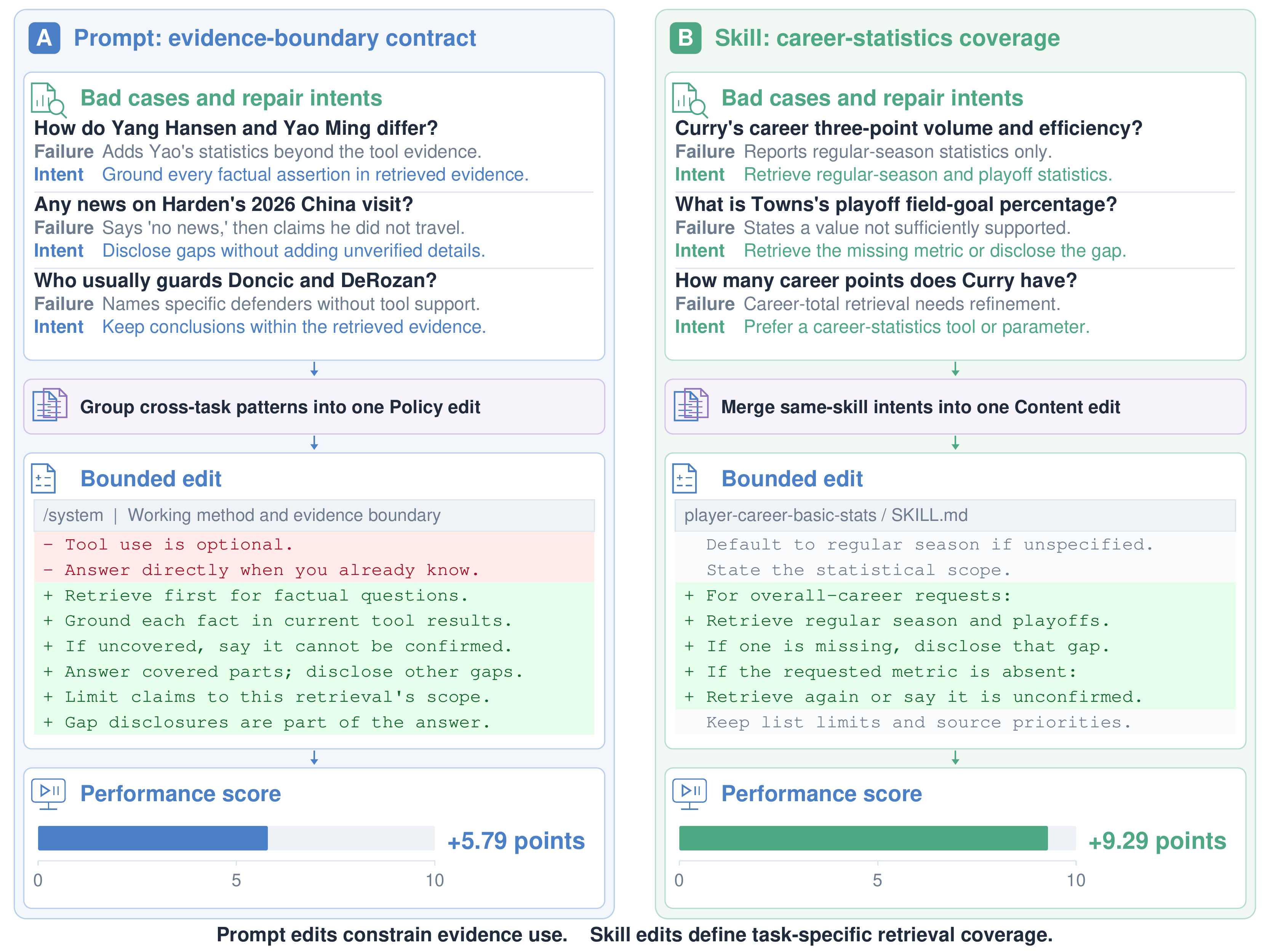}
  \caption{From concrete failures to layer-specific repairs. Related repair intents lead to a bounded Policy or Content edit. In the diff panels, red denotes removed text, green added text, and gray retained context. Performance score bars show mean gains on the respective held-out evaluation sets. The cases come from separate experiments; their gains are not a controlled comparison or estimates of an individual rule's contribution.}
  \label{fig:harness-case-study}
\end{figure}

We examine two offline evolution events to illustrate how failure evidence leads to edits at different harness layers. Case A repairs a cross-task evidence-boundary gap through a prompt-level Policy edit; Case B extends career-statistics retrieval through a skill-level Content edit. Figure~\ref{fig:harness-case-study} connects representative questions and repair intents to the resulting edits and held-out performance score gains. 

\subsubsection{Case A: A Cross-Task Evidence-Boundary Contract}
\label{sec:harness-case-policy}

\paragraph{Failure Evidence and Repair Intents.}
The baseline system prompt allowed the agent to answer without retrieval when it believed it already knew the answer. Three representative failures exposed the resulting evidence-boundary gap. A comparison of Yang Hansen and Yao Ming added Yao's statistics beyond the retrieved evidence. A response about James Harden's 2026 China visit first stated that no public information was available, then asserted that he had not traveled with the team. A response about defensive matchups involving Luka Don\v{c}i\'c and DeMar DeRozan named specific defenders without tool support.

These failures motivated related repair intents: ground factual assertions in current tool evidence, disclose missing information without adding unverified details, and keep conclusions within the retrieval scope. Diagnosis grouped the cross-task patterns around one global evidence-boundary contract. Because the same root cause appeared across question types, the selected target was the system prompt rather than individual skills.

\paragraph{Bounded Policy Edit.}
The candidate replaced the permissive retrieval instruction with a coordinated contract: retrieve before answering factual questions, support verifiable assertions with current tool results, and state uncertainty when the requested information is not covered. For partially covered questions, the response should answer the supported portion and disclose the remaining gap. Gap disclosures were explicitly treated as part of the answer, keeping the evidence requirement consistent with existing concision rules.

The edit was confined to the relevant system-prompt section and introduced no player-specific facts or answer templates. Independent audit and validation checked its scope and consistency. For the second (player-visit) example, the revised policy requires the agent to disclose that the retrieved evidence does not establish whether the player traveled with the team, rather than infer that he did not. This preserves the distinction between missing evidence and evidence supporting a negative conclusion.

\paragraph{Held-Out Performance.}
On the held-out evaluation set, the mean Performance score increased from 65.31 to 71.10, a gain of 5.79 points relative to the historical baseline. The revised prompt was accepted. This result reflects the overall contract update rather than the isolated contribution of any one clause.

\subsubsection{Case B: Task-Specific Career-Statistics Coverage}
\label{sec:harness-case-content}

\paragraph{Failure Evidence and Repair Intents.}
The skill-only experiment started with a prompt that already contained Case A's evidence contract. A question about Stephen Curry's career three-point volume and efficiency activated \nolinkurl{player-career-basic-stats}, but the agent retrieved only regular-season statistics. The returned values were grounded in tool results; the gap was that the evaluation convention for this overall-career request also required playoff coverage, which the answer neither supplied nor identified as missing.

The resulting intent called for retrieving both season types and disclosing any missing portion. Two related cases supported a broader repair of the same skill: a question about Karl-Anthony Towns's playoff field-goal percentage called for retrieving the requested metric or stating that it could not be confirmed, while a historical question about Curry's career points called for a more appropriate career-statistics retrieval route. These intents accumulated across events and were grouped by their shared target skill. The deficiency concerned task-specific retrieval knowledge, so it was attributed to Skill Content rather than to the global evidence policy.

\paragraph{Bounded Content Edit.}
The candidate added two requirements: retrieve both regular-season and playoff statistics for an explicit overall-career request, and retrieve again or disclose the gap when the initial result lacks the requested metric. The existing default to regular-season statistics for unspecified requests, tool-covered list boundaries, and preferred data sources were preserved. Only the target skill changed; the system prompt remained fixed. The overall-career convention was specific to this task family and did not redefine the scope of all sports-statistics questions.

Replay of the Curry question showed the corresponding execution change. The baseline query selected regular-season data only; the revised query selected both regular-season and playoff data and grouped the results by season type. The answer then reported the two categories separately. This trajectory shows how a content edit expanded retrieval and answer coverage while preserving the original supported statistics.

\paragraph{Held-Out Performance.}
On the relevant held-out evaluation subset, the skill update improved the mean performance score by 9.29 points and was accepted. This aggregate result covers the target skill's update; the query trajectory provides the direct evidence for the season-coverage change.

\subsubsection{Layer Responsibilities}
\label{sec:harness-case-discussion}

The cases share a path from failure evidence through repair intents to bounded edits and validation, but address different responsibilities. The prompt constrains which factual assertions may appear in an answer; the skill specifies which data and statistical scope a task requires. Case B occurred with Case A's contract already present, showing that grounded assertions alone do not guarantee complete task coverage. Placing specialized retrieval rules in the prompt would blur this boundary, while repeating global evidence rules in individual skills would duplicate policy and make its application depend on skill activation. Effective repair therefore requires routing each diagnosed gap to the layer with the appropriate editing authority.

%% file: Sports_Agent_Anonymization/AutoEval.tex
\section{A Unified Feedback Engine for Self-Evolution}
\label{sec:eval}

Agent tasks typically involve reasoning, tool use, and information synthesis. Assessing answer correctness alone is generally insufficient to explain how a failure arose or identify the component that requires improvement~\citep{zhang2025whoandwhen}. Actionable feedback for self-evolution therefore requires execution evidence to be examined in relation to task requirements, so that an observed failure can be linked to a specific recommendation for improvement.

The Feedback Engine comprises two modules: Assessment and Diagnosis. Assessment evaluates task performance under a common specification~\citep{zheng2023llmjudge}. For confirmed factual errors, Diagnosis investigates the execution evidence to identify the mechanisms and components involved. The resulting feedback specifies the optimization targets, update unit, diagnostic conclusion, and recommended improvement.

Section~\ref{subsec:autoeval-formulation} formalizes the process from runtime evidence to evolution feedback. Section~\ref{subsec:autoeval-methods} presents the methods used by Assessment and Diagnosis, and Section~\ref{subsec:autoeval-cases} examines two production cases that inform recommendations for model training and harness refinement.

\subsection{From Runtime Evidence to Evolution Signals}
\label{subsec:autoeval-formulation}

Following Section~\ref{sec:overview}, the Feedback Engine operates on the runtime-evidence representation $e_t = (q_t, y_t, \tau_t)$, which comprises the input query, final answer, and task-level execution trace. Assessment and Diagnosis use only the fields relevant to their respective functions. The trace captures observable execution information, including planning decisions, applicable instructions, tool calls and results, and model responses. An evaluation specification $S$ defines each metric's scope, applicability conditions, and evidence requirements; the associated rubrics describe observable performance at different quality levels~\citep{liu2023geval}. All tasks in an evaluation batch use the same specification. Assessment returns

\begin{equation}
\mathcal A_t=\operatorname{Assess}(e_t;S)
=\{A_{t,m}:m\in S\}.
\end{equation}

The set \(\mathcal A_t\) contains the assessment results. For each metric \(m\), \(A_{t,m}\) records the evaluation status, rationale, evidence references, and score, where available. These results identify unmet quality requirements, but explaining their causes requires further investigation.

For a confirmed factual error, let \(\phi_i\) denote the affected claim and \(r_i\) the factual reference, comprising verified facts, their scope, and their sources. After establishing this reference, the request is resubmitted in new sessions to test whether the error recurs and to obtain additional execution evidence. The reference remains fixed throughout the investigation. Let \(\tau_t^{(j)}\) denote the trace available from run \(j\). Diagnosis produces

\begin{equation}
\mathcal D^{\mathrm{diag}}_{t,i,j}=\operatorname{Diagnose}
\left(q_t,\phi_i,r_i,\tau_t^{(j)}\right).
\end{equation}

The result \(\mathcal D^{\mathrm{diag}}_{t,i,j}\) records the relevant components, error mechanism, expected behavior, and unresolved questions for claim \(i\) in run \(j\). The corresponding results are collected in \(\mathcal D_t^{\mathrm{diag}}\). Fact verification and controlled reproduction supplement the initial runtime evidence with reference facts and replay observations. Only cases requiring further analysis proceed to root-cause diagnosis, as similar factual errors may arise from different failure mechanisms. The resulting structured case records can be aggregated into recurring capability gaps and used to guide targeted training-data construction.

Assessment and Diagnosis jointly produce actionable recommendations for system improvement. We represent each feedback item as follows:

\begin{equation}
\gamma=\langle\mathcal Z^{\mathrm{opt}},u,d,\mathrm{rec}\rangle.
\end{equation}

The set \(\mathcal Z^{\mathrm{opt}}\) specifies the optimization targets, and \(u\) identifies the component to be modified, such as a Prompt or Skill. The diagnostic conclusion \(d\) records the error mechanism, supporting evidence, scope, and remaining uncertainty. The recommendation \(\mathrm{rec}\) states the capability, rule, or behavior to improve. For feedback directed at the model, the update unit need not be specified at the level of an internal model component.

The stage at which an error appears does not, by itself, determine the component to update. An error in query construction may reflect failure to follow an available rule or a gap in the task guidance. The former may motivate model training or execution constraints, whereas the latter may require a Skill revision; both changes may also be warranted. The recommendation must therefore remain grounded in the diagnostic report, with evidence references and conclusion status retained in \(d\).

For Sports Agent, the optimization targets are the Harness and the Agentic SQL Model. Let \(\Gamma_t\) denote the set of evidence-supported feedback items for which an improvement direction can be determined, and let \(\mathcal Z^{\mathrm{opt}}(\gamma)\) denote the target set of item \(\gamma\). Using \(\mathrm H\) and \(\mathrm M\) for the harness and model, respectively, we define the target-specific evolution signals as

\begin{equation}
\begin{aligned}
g_t^H&=\{\gamma\in\Gamma_t:\mathrm H\in\mathcal Z^{\mathrm{opt}}(\gamma)\},\\
g_t^M&=\{\gamma\in\Gamma_t:\mathrm M\in\mathcal Z^{\mathrm{opt}}(\gamma)\}.
\end{aligned}
\end{equation}

A feedback item may belong to both sets, and either set may be empty. When a diagnosis supports distinct recommendations for different components, the recommendations can be represented as separate items sharing the same diagnostic evidence. Conversely, findings that do not yet justify a specific update remain in the investigation record. The resulting feedback is used by the model and harness update procedures described in Sections~\ref{sec:model-loop} and~\ref{sec:harness-loop}, respectively.

Table~\ref{tab:autoeval-feedback-example} illustrates the feedback representation using the leaderboard eligibility case discussed in Section~\ref{subsubsec:eligibility-case}.

\begin{table}[htbp]
\small
\centering
\caption{Fields and example values of structured improvement feedback.}
\label{tab:autoeval-feedback-example}
\linespread{1.10}\selectfont
\renewcommand{\arraystretch}{1.35}
\begin{tabularx}{\linewidth}{
  @{}
  >{\hsize=0.48\hsize\linewidth=\hsize\raggedright\arraybackslash}X
  >{\hsize=1.52\hsize\linewidth=\hsize\raggedright\arraybackslash}X
  @{}}
\toprule
\textbf{Feedback field}
  & \textbf{Example: leaderboard eligibility} \\
\midrule[0.9pt]
Optimization targets \(\mathcal Z^{\mathrm{opt}}\)
  & \(\{\mathrm H\}\): Harness \\
\arrayrulecolor{black!20}
\midrule[0.3pt]
\arrayrulecolor{black}
Update unit \(u\)
  & Leaderboard Skill \\
\arrayrulecolor{black!20}
\midrule[0.3pt]
\arrayrulecolor{black}
Diagnosis \(d\)
  & The SQL eligibility criterion used field-goal attempts (FGA) rather than field goals made (FGM). Supplementary inspection of the same request established that the loaded Skill and table documentation omitted the leaderboard's eligibility requirement \\
\arrayrulecolor{black!20}
\midrule[0.3pt]
\arrayrulecolor{black}
Recommendation \(\mathrm{rec}\)
  & Specify the standard eligibility criterion and its source in the leaderboard Skill, retain user-defined conditions, and validate the distinction between makes and attempts and the eligibility boundaries \\
\bottomrule
\end{tabularx}
\end{table}

\begin{figure}[htbp]
\centering
\includegraphics[width=\linewidth]{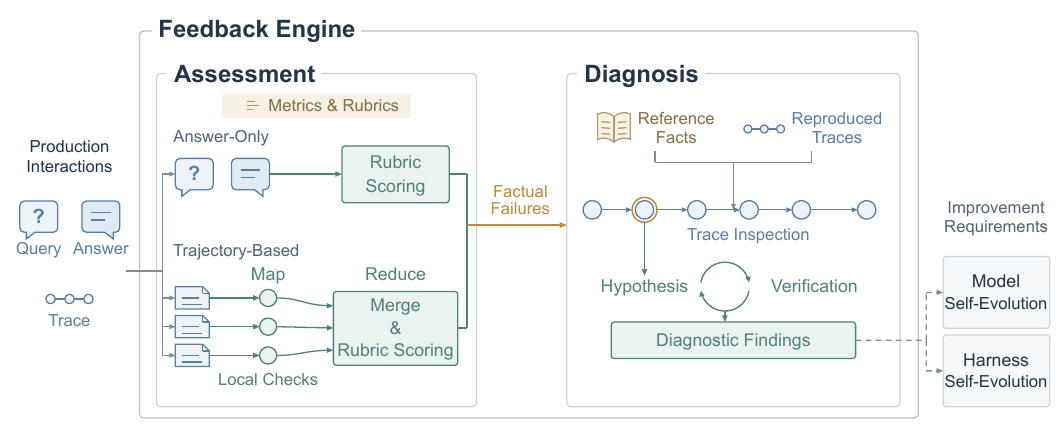}
\caption{Assessment applies a shared evaluation specification through direct rubric-based scoring or trace-based Map--Reduce~\citep{dean2004mapreduce}, with Map performing local verification and Reduce aggregating evidence and assigning scores. For confirmed factual errors, Diagnosis uses factual references and reproduced traces to investigate hypotheses and produce findings that inform recommendations for the model and harness.}
\label{fig:autoeval}
\end{figure}

Producing evidence-based recommendations therefore requires dispersed records to be integrated and interpreted in relation to the observed execution failure. Figure~\ref{fig:autoeval} summarizes how Assessment and Diagnosis connect evidence analysis to recommendations for improvement.

\subsection{Generating Feedback for Self-Evolution}
\label{subsec:autoeval-methods}

Section~\ref{subsec:autoeval-formulation} defines how evaluation and diagnostic findings are represented as improvement feedback. Selecting an appropriate update, however, requires both a judgment of task performance and an explanation of the observed failure. Assessment identifies unmet task requirements, while Diagnosis investigates the mechanisms underlying confirmed factual errors. Their findings jointly support the choice of optimization target, update unit, and recommended change.

\subsubsection{Evidence-Based Quality Assessment}
\label{subsubsec:evojudge}

Evaluation metrics require different forms of evidence. Answer relevance can be assessed directly from the question and response, whereas factual-claim verification and behavioral evaluation require evidence from the execution trace~\citep{zhuge2025agentjudge}. Combining evidence retrieval, claim--evidence verification, and scoring in a single model call can reduce judgment reliability, particularly when the call must process a long trace and handle several subtasks simultaneously. Assessment therefore selects an execution path according to the metric specification and separates local verification from final scoring when trace evidence is required. Table~\ref{tab:evojudge-paths} summarizes these paths.

For claim checking, Assessment extracts factual claims \(\Phi_t=\{\phi_i\}\) from \((q_t,y_t)\)~\citep{min2023factscore,ru2024ragchecker} and distinguishes core from secondary claims. Each evidence unit \(z_\ell\) corresponds to an identifiable trace record containing its source type, text, and reference. Context checking instead examines metric-relevant information and behavior without extracting claims. Successful execution under the selected mode is represented by

\begin{equation}
\begin{aligned}
\mathcal O_{t,m}^{(\ell)}
&=
\begin{cases}
\operatorname{Map}(q_t,y_t,z_\ell,\Phi_t;m),
&\text{claim checking},\\
\operatorname{Map}(q_t,y_t,z_\ell;m),
&\text{context checking},
\end{cases}\\
\mathcal O_{t,m}
&=\operatorname{Collect}\!\left(
[\mathcal O_{t,m}^{(\ell)}]_\ell
\right).
\end{aligned}
\end{equation}

Each \(\mathcal O_{t,m}^{(\ell)}\) is a validated list of local observations linked to their sources and may contain zero or more items recording support, contradiction, qualification, or context. The \(\operatorname{Collect}\) operation aggregates these lists in evidence-unit order and removes duplicates to obtain \(\mathcal O_{t,m}\). Map verifies evidence relationships under metric \(m\) without assigning scores. Tool outputs and instruction snapshots can substantiate claims within the recorded evidence; the agent's own statements are retained only as context.

\begin{table}[htbp]
    \small
    \centering
    \caption{Evaluation targets, evidence requirements, and execution paths.}
    \label{tab:evojudge-paths}
    \linespread{1.10}\selectfont
    \renewcommand{\arraystretch}{1.35}
    \begin{tabularx}{\linewidth}{
      @{}
      >{\hsize=0.78\hsize\linewidth=\hsize\raggedright\arraybackslash}X
      >{\hsize=0.95\hsize\linewidth=\hsize\raggedright\arraybackslash}X
      >{\hsize=1.27\hsize\linewidth=\hsize\raggedright\arraybackslash}X
      @{}}
    \toprule
    \textbf{Evaluation target}
      & \textbf{Required evidence}
      & \textbf{Execution path} \\
    \midrule[0.9pt]
    Response quality
      & Question and answer
      & Direct rubric-based scoring \\
    \arrayrulecolor{black!20}
    \midrule[0.3pt]
    \arrayrulecolor{black}
    Claim--evidence relationships
      & Claims in the answer and relevant trace evidence
      & Local verification followed by scoring in Reduce \\
    \arrayrulecolor{black!20}
    \midrule[0.3pt]
    \arrayrulecolor{black}
    Behavior in context
      & Information available and actions taken during execution
      & Criterion-specific observations followed by scoring in Reduce \\
    \bottomrule
    \end{tabularx}
\end{table}

Once the local observations have been collected, the claim-checking path proceeds to Reduce. The complete claim list is used to consolidate verification states, which in turn constrain rubric-based scoring:

\begin{equation}
\begin{aligned}
\boldsymbol s_{t,m}
&=\operatorname{BuildStates}(\Phi_t,\mathcal O_{t,m}),\\
A_{t,m}
&=\operatorname{Score}_m
(q_t,y_t,\boldsymbol s_{t,m},\mathcal O^{\mathrm{ctx}}_{t,m}).
\end{aligned}
\end{equation}

The sequence \(\boldsymbol s_{t,m}\) records each claim's content, importance, verification status, and evidence references, including claims without relevant observations. Contextual observations that are not associated with a specific claim are retained in \(\mathcal O^{\mathrm{ctx}}_{t,m}\). Given these records, \(\operatorname{Score}_m\) produces a rubric-based evaluation subject to the claim-state constraints: the model forms an initial assessment within the permitted score range, and structurally valid outputs undergo programmatic checks and any necessary corrections to the score, status, and rationale. Context checking evaluates the collected observations directly against the rubric, without constructing claim states or applying these constraints. Both paths retain rationales and evidence references without rereading the full trace. Support within a recorded trace remains distinct from independent factual verification.

\subsubsection{From Evaluation Findings to Component Diagnosis}
\label{subsubsec:rca}

The same factual error may have different causes, and reproduction alone does not identify the components or behaviors responsible. Diagnosis therefore organizes evidence collection and verification around explicit hypotheses, using the findings to refine explanations and inform recommendations~\citep{zhu2025agentdebug}. Once the factual reference \(r_i\) has been established, the request is repeated in new sessions. With \(y_t^{(j)}\) denoting the answer from run \(j\), the recurrence condition and claim--request pairs eligible for investigation are

\begin{equation}
\begin{aligned}
R_{ij}&=\mathbf1\!\left[
\operatorname{Related}(y_t^{(j)},\phi_i)
\land\operatorname{Conflicts}(y_t^{(j)},r_i)
\right],\\
\mathcal I_t^{\mathrm{diag}}&=\{(i,j):R_{ij}=1,\ \tau_t^{(j)}\text{ is available}\}.
\end{aligned}
\end{equation}

The condition \(R_{ij}=1\) holds when the new answer contains a claim semantically related to \(\phi_i\) that still conflicts with \(r_i\); identical wording or the same erroneous value is not required. Each pair in \(\mathcal I_t^{\mathrm{diag}}\) links a claim to a particular run. Because session conditions may differ from those of the original run, reproduction is not a deterministic replay. Diagnosis examines the observed inputs, actions, and outputs and traces upstream within the limits of the available evidence, distinguishing missing or unloaded capabilities, inadequate task guidance, and failure to follow an available rule.

The verification method depends on the hypothesized mechanism. Instruction checks first establish whether the relevant rule entered the request context before the action under investigation, then compare the rule with subsequent behavior. Answer consistency checks compare the affected claims with the original tool outputs. SQL-related hypotheses can be examined through read-only local comparisons under the preparation, registration, and verification requirements in Table~\ref{tab:rca-sql-contrast}.

\begin{table}[htbp]
\small
\centering
\caption{Requirements for read-only local SQL comparisons.}
\label{tab:rca-sql-contrast}
\linespread{1.10}\selectfont
\renewcommand{\arraystretch}{1.35}
\begin{tabularx}{\linewidth}{
  @{}
  >{\hsize=0.30\hsize\linewidth=\hsize\raggedright\arraybackslash}X
  >{\hsize=1.70\hsize\linewidth=\hsize\raggedright\arraybackslash}X
  @{}}
\toprule
\textbf{Comparison stage}
  & \textbf{Requirement} \\
\midrule[0.9pt]
Preparation
  & Inspect the original tool output before registering a hypothesis, and inspect the relevant table documentation before executing a query \\
\arrayrulecolor{black!20}
\midrule[0.3pt]
\arrayrulecolor{black}
Registration
  & Register the predicted values, entities and statistical scope to be held fixed, proposed query modification, and output columns for both computations \\
\arrayrulecolor{black!20}
\midrule[0.3pt]
\arrayrulecolor{black}
Verification
  & Execute both computations in a single read-only query, compare the registered output columns with the predictions, and record warnings and unexplained discrepancies \\
\bottomrule
\end{tabularx}
\end{table}

The tools enforce these prerequisites so that each comparison is grounded in the original output and documented data semantics. Predictions for the baseline and modified queries are registered before execution. Verification then checks whether the baseline reproduces the original erroneous value and whether the modified query yields the predicted target.

A comparison is accepted as verification evidence only if it passes the applicable checks without warnings. Otherwise, unexplained discrepancies require revision of the hypothesis or further upstream investigation. Even a successful comparison provides only local support for the proposed error mechanism; the programmatic checks do not establish that the query modification isolates a single factor affecting the result.

Throughout an investigation, the system records the evidence examined, open hypotheses, verification conclusions, and unexplained discrepancies, and identifies outstanding tasks. Before delivery, it checks that the conclusions and tests have been recorded and linked to runtime evidence. Insufficient support requires further investigation or an explicit statement of what remains unresolved. The labels confirmed, hypothesis, and inconclusive describe evidential support rather than calibrated probabilities.

The distinction between missing guidance and noncompliance also determines the type of recommendation. If an applicable rule was present in the execution context but was not followed, training or execution constraints may improve adherence. If the guidance omitted a necessary rule, the appropriate target may instead be a Prompt or Skill. Defects in implementation, configuration, or knowledge and data dependencies may indicate other changes. Accordingly, the diagnosis \(d\), update unit \(u\), and recommendation \(\mathrm{rec}\) must be supported by the same body of evidence.

\subsection{Case Studies: From Observed Failures to Evolution Feedback}
\label{subsec:autoeval-cases}

We present two representative cases with different improvement targets. Assessment first verified factual accuracy using multiple online sources, after which Diagnosis analyzed confirmed errors across repeated executions. The figures summarize the case comparisons and distinguish automated findings from supplementary evidence supporting the recommendations.

\subsubsection{SQL Tool Diagnosis for Model Self-Evolution}
\label{subsubsec:appearance-case}

In a representative evaluation case, the system reported that Kobe Bryant had 920 career regular-season wins. External verification established the correct count as 836 games in which he appeared and his team won, while repeated controlled executions continued to return 920.

\begin{figure}[htbp]
    \centering
    \includegraphics[width=0.8\linewidth]{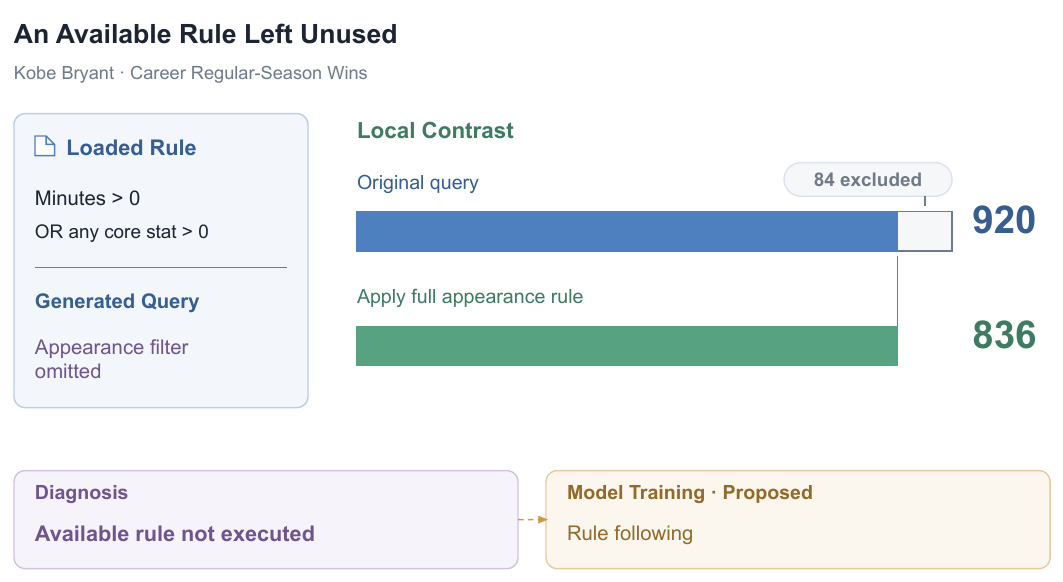}
    \caption{Archived comparison for Kobe Bryant's career regular-season wins and the proposed model-training recommendation.}
    \label{fig:appearance-case}
\end{figure}

Trace inspection showed that the query applied the player, regular-season, and win filters but omitted the appearance condition specified in the pre-query guidance: positive minutes played or a positive value in at least one core statistic. In the archived comparison, applying this condition reduced the count from 920 to 836; the 84 excluded records fully explained the discrepancy (Figure~\ref{fig:appearance-case}). The final answer matched the original query result.

The evidence supports a diagnosis of failure to follow an available rule. We therefore recommend constructing SQL training examples for the Agentic SQL Model that reinforce application of the appearance rule, with validation across players and appearance conditions.

\subsubsection{Skill Refinement for Harness Self-Evolution}
\label{subsubsec:eligibility-case}

The standard eligibility criterion for the career regular-season field-goal percentage leaderboard is at least 2,000 field goals made (FGM). The question specified neither statistical scope nor eligibility criteria, but the response applied a threshold of 2,000 field-goal attempts (FGA), ranking Daniel Gafford first at approximately 70.2\%.

\begin{figure}[htbp]
    \centering
    \includegraphics[width=0.8\linewidth]{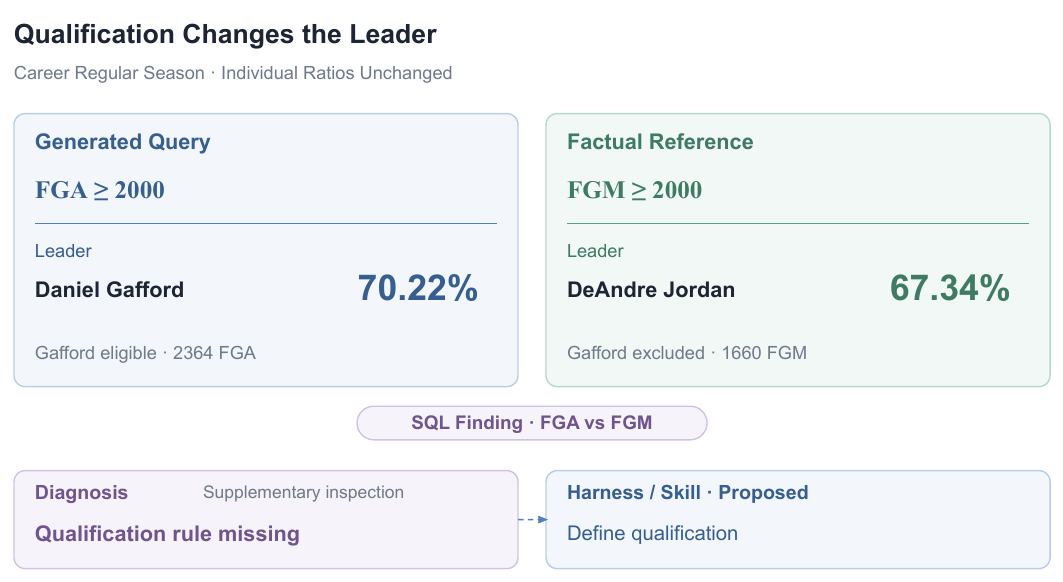}
    \caption{Comparison of eligibility criteria for the career field-goal percentage leaderboard and the proposed Skill refinement.}
    \label{fig:eligibility-case}
\end{figure}

The query correctly computed career FGM divided by FGA; Gafford's 1,660 makes in 2,364 attempts yielded 70.22\%. The archived comparison retained the statistical scope and ratio calculation but changed eligibility to FGM\(\ge\)2000, excluding Gafford and placing DeAndre Jordan first at 67.34\% (Figure~\ref{fig:eligibility-case}). The set of eligible players changed; Gafford's percentage did not.

Diagnosis confirmed the error in the SQL eligibility field. The recommendation to revise the Skill is additionally supported by inspection of the same request: the guidance loaded before query generation specified the statistical scope and ratio calculation but omitted the leaderboard's eligibility requirement. The proposed revision is to document the standard criterion and its source, preserve user-defined conditions, and validate the distinction between makes and attempts and the eligibility boundaries.

Across these cases, the Feedback Engine combines quality assessment, component diagnosis, and task-rule evidence to produce specific recommendations for improvement. The resulting feedback can inform model learning objectives and the development of reusable Skill specifications.

%% file: Sports_Agent_Anonymization/Evaluation.tex
\section{Production Deployment and Evaluation}
\label{sec:evaluation}

\subsection{Production System Overview}
\label{sec:nba-chat}
The evaluated system is a conversational sports analytics product powered by specialized models and an agent system. Its first production version was launched in June 2026, serving users in mainland China. The product integrates structured databases, knowledge bases, and real-time web sources, enabling users to retrieve and analyze sports information without manually navigating disparate data platforms. It supports knowledge retrieval, statistical analysis, real-time information access, player and team profiling, pre-game analysis, and contextual multi-turn follow-up. Users can explore historical records, compare player and team performance, aggregate statistics, track schedules, results, standings, injuries, and transactions, and discuss potential game outcomes using available evidence. A unified conversational interface allows users to refine questions, adjust analytical constraints, and continue exploring related topics across multiple turns.

To address the limitations of static configurations in a changing production environment, we upgraded the baseline system between June and September 2026 by integrating the Sports Agent framework. The evolved system combines an improved Agentic SQL Model with an updated Sports Agent Harness, supported by the Feedback Engine for runtime assessment, failure diagnosis, and update validation. The upgrade targeted three objectives: improving factual correctness, enhancing response richness and informativeness, and reducing end-to-end latency. The following subsections evaluate these objectives through comparisons between the baseline and evolved systems, continuous-evolution experiments, production traffic analysis, and user engagement and retention statistics.

To protect user privacy, all workload and engagement statistics reported in this section are derived from aggregated, de-identified product data. The underlying analyses exclude raw user content and persistent user identifiers. Additional details on data scope and privacy safeguards are provided in Appendix~\ref{app:data-privacy}.

\subsection{Overall Performance: Baseline vs. Evolved System}
We compare the baseline and evolved systems on a shared evaluation set. The baseline uses a largely static agent configuration, whereas the evolved system integrates an evolved Agentic SQL Model and Sports Agent Harness, supported by the Feedback Engine for runtime assessment, error attribution, and version validation. The evaluation set contains 500 questions: 400 internally curated evaluation samples and 100 questions generated by an AI agent based on daily trending topics to represent emerging information needs. It covers factual retrieval, statistical analysis, real-time information, and composite queries. Response quality is assessed using a multidimensional evaluation framework covering factual correctness and content richness.

\begin{figure}[htbp]
  \centering
  \includegraphics[width=\linewidth]{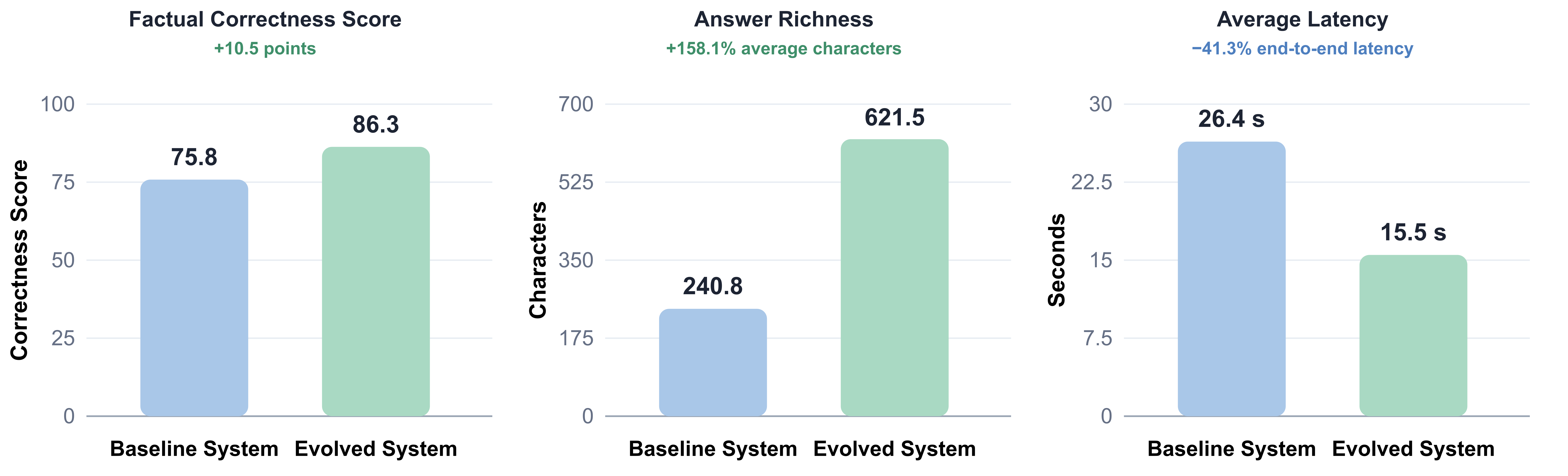}
  \caption{Overall performance comparison between the baseline and evolved systems.}
  \label{fig:nba-chat-performance}
\end{figure}

We report three user-facing metrics. \textbf{Factual Correctness Score} is an LLM-as-a-Judge assessment of the correctness of core factual claims, normalized to a 0-100 scale. \textbf{Answer Richness} uses average response length as an observable proxy for informational breadth and completeness of the response. \textbf{Average Latency} measures the end-to-end time from request receipt to response completion. As shown in Figure~\ref{fig:nba-chat-performance}, the evolved system achieves substantial improvements in response quality and execution efficiency. The factual correctness score rises from 75.8 to 86.3, a gain of 10.5 points, while average response length increases from 240.8 to 621.5 characters, suggesting more informative answers. Meanwhile, average end-to-end latency falls from 26.4 to 15.5 seconds, a reduction of 41.3\%. These results indicate that coordinated model and harness evolution can improve factual correctness and response richness while reducing latency, rather than trading response efficiency for answer quality.

\subsection{Continuous Evolution in Production}
To assess sustained improvement in a deployed setting, we evaluated 30 successive harness versions produced through iterative evolution. Each update builds on the preceding version and uses relevant execution evidence and Feedback Engine diagnostics to identify capability gaps and refine the prompt configuration and associated skills. All versions, V1-V30, are evaluated on the same fixed test set under a consistent evaluation protocol.

The aggregate Performance Score combines general response quality and application-specific criteria, weighted at 30\% and 70\%, respectively. General criteria include readability, self-consistency, and relevance; application-specific criteria comprise expressiveness, factual correctness, and faithfulness to retrieved evidence. This fixed-set evaluation tracks improvements driven by newly observed failures while monitoring regressions in established capabilities.

\begin{figure}[htbp]
  \centering
  \includegraphics[width=\linewidth]{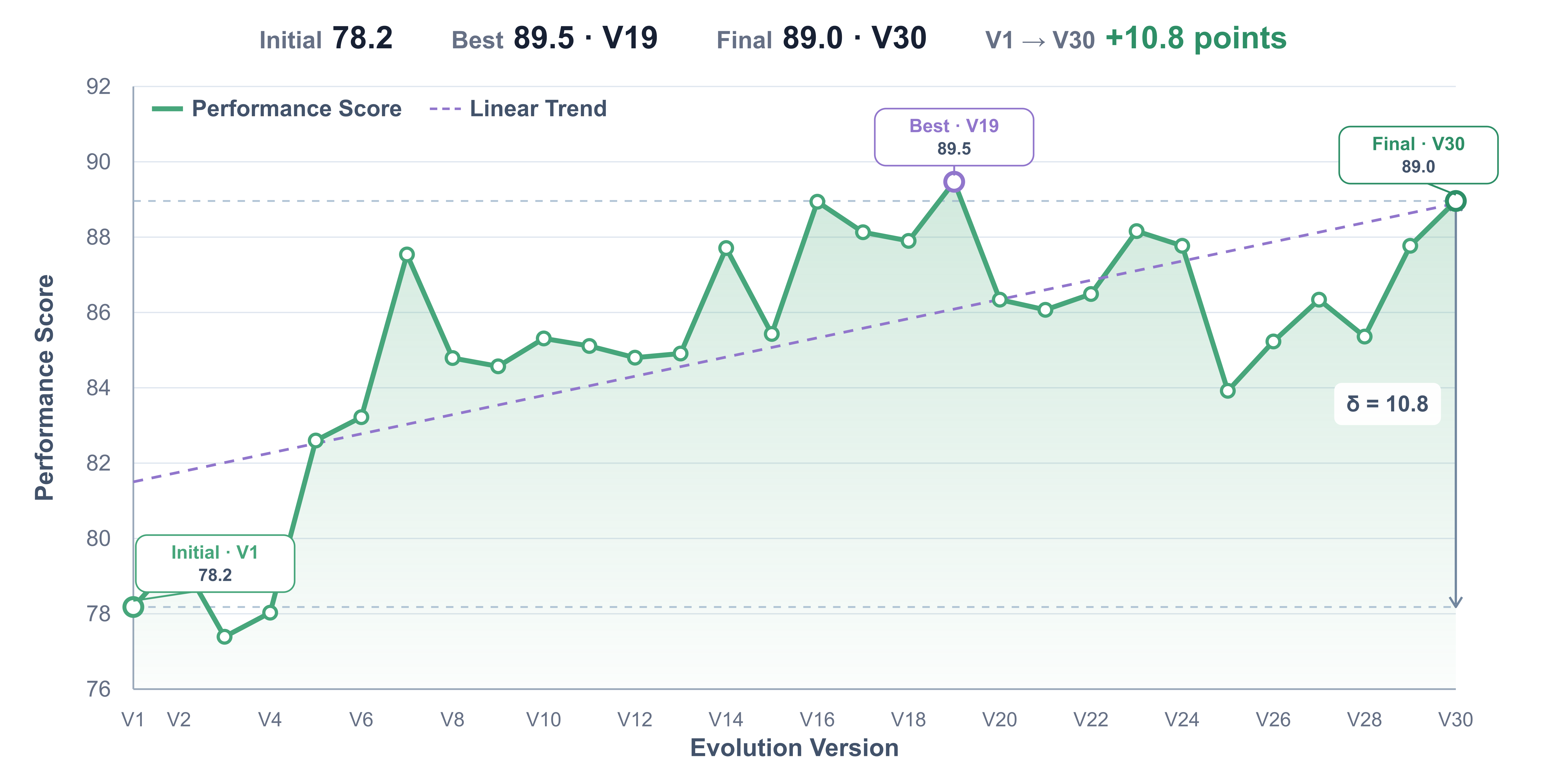}
  \caption{Performance over 30 self-improvement iterations. The aggregate score increases from 78.2 at V1 to 89.0 at V30 and reaches its maximum of 89.5 at V19.}
  \label{fig:evolution-performance}
\end{figure}

As shown in Figure~\ref{fig:evolution-performance}, the Performance Score increases from 78.2 at V1 to 89.0 at V30, a gain of 10.8 points, and peaks at 89.5 at V19. Despite fluctuations across individual versions, the fitted trend is positive, indicating overall improvement across evolution rounds. The final version remains within 0.5 points of the peak score. However, local declines highlight the non-monotonic nature of harness evolution: updates targeting one failure category may adversely affect other tasks. Fixed-set regression evaluation and release gating are therefore essential at each round.

\subsection{Production Workload Profile}

% \subsubsection{Deployment Scale}
% From its launch on June 5 through November 3, 2026, the product processed hundreds of thousands of user requests, averaging several thousand on game days and reaching a daily peak of over ten thousand. The launch coincided with the league finals, when heightened event attention and overall activity in the host application contributed to an immediate traffic peak. Request volume declined and stabilized during the subsequent offseason, then rose again as the preseason and regular season began in October. The seven-day moving average shows the same pattern. These observations suggest that product traffic is closely associated with the competition calendar and broader application activity. The system must therefore accommodate peak demand around major events while sustaining service as information needs change across competition phases.

% \begin{figure}[htbp]
%   \centering
%   \includegraphics[width=\linewidth]{figs/Daily-User-Request-Traffic.png}
%   \caption{Relative daily request volume and its seven-day moving average across competition phases.}
%   \label{fig:user-request-traffic}
% \end{figure}

\subsubsection{Query Distribution}
The query distribution varies across seasonal phases in topic composition, concentration, and tool requirements. Player information consistently accounts for 27.0\%--31.4\% of queries, while transaction-related queries rise sharply from 3.4\% during the playoffs to 32.1\% in the offseason. Together, these two categories represent 59.1\% of offseason queries, reflecting a more concentrated distribution. As the regular season begins, the combined share of game prediction, schedule, and game-analysis queries increases from 8.0\% to 20.4\%, redirecting attention toward ongoing competition and statistical analysis. These shifts also reshape tool requirements. In-season queries rely more heavily on structured databases, whereas offseason queries require greater use of news search and knowledge retrieval. Adapting to production query shifts therefore requires corresponding adjustments to task planning, source selection, and tool orchestration.

\begin{figure}[htbp]
  \centering
  \includegraphics[width=\linewidth]{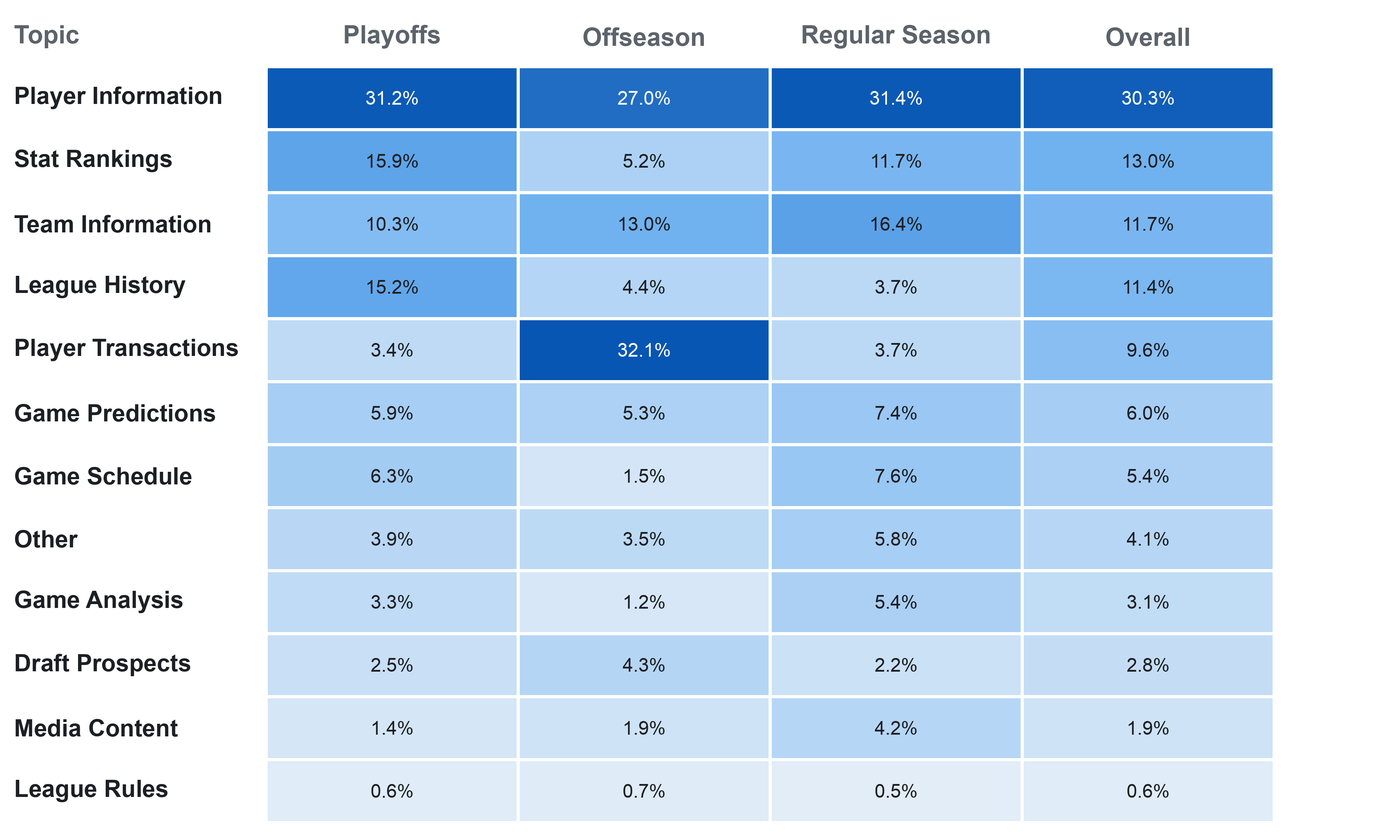}
  \caption{Query-topic distribution across the playoffs, offseason, and regular season.}
  \label{fig:nba-topic-heatmap}
\end{figure}

\subsubsection{Tool and Complexity Profile}
Tool use is shaped jointly by the season phase and the deployed system version. During the offseason, input queries center on trades, drafts, and roster changes: Web Search and Profile RAG achieve coverage rates of 57.0\% and 36.5\%, respectively, while Database Query drops to 17.0\%, reflecting the dominance of current news and unstructured knowledge.  During the regular season, queries shift back toward schedules, box scores, and team performance. Consequently, Database Query coverage rises to 40.6\%, nearly matching Web Search at 41.3\%.

The regular-season observation period coincided with the deployment of the evolved system, which improved context management, intermediate-evidence reuse, and task-termination policies, thereby reducing redundant retrieval during follow-up reasoning. During the playoffs and offseason, 73.3\% and 89.1\% of requests required at least two tool calls, respectively. After the deployment of the evolved system, this proportion fell to 33.6\%, with single-call requests rising to 46.1\% and three-or-more-call requests dropping to 12.9\%. Combined with the gains in answer quality, these results indicate that fewer tool calls reflect better context utilization and more compact execution paths rather than reduced capability. In short, the evolved agent system preserves multi-source evidence acquisition while substantially lowering the interaction cost per task.

\begin{figure}[htbp]
  \centering
  \includegraphics[width=\linewidth]{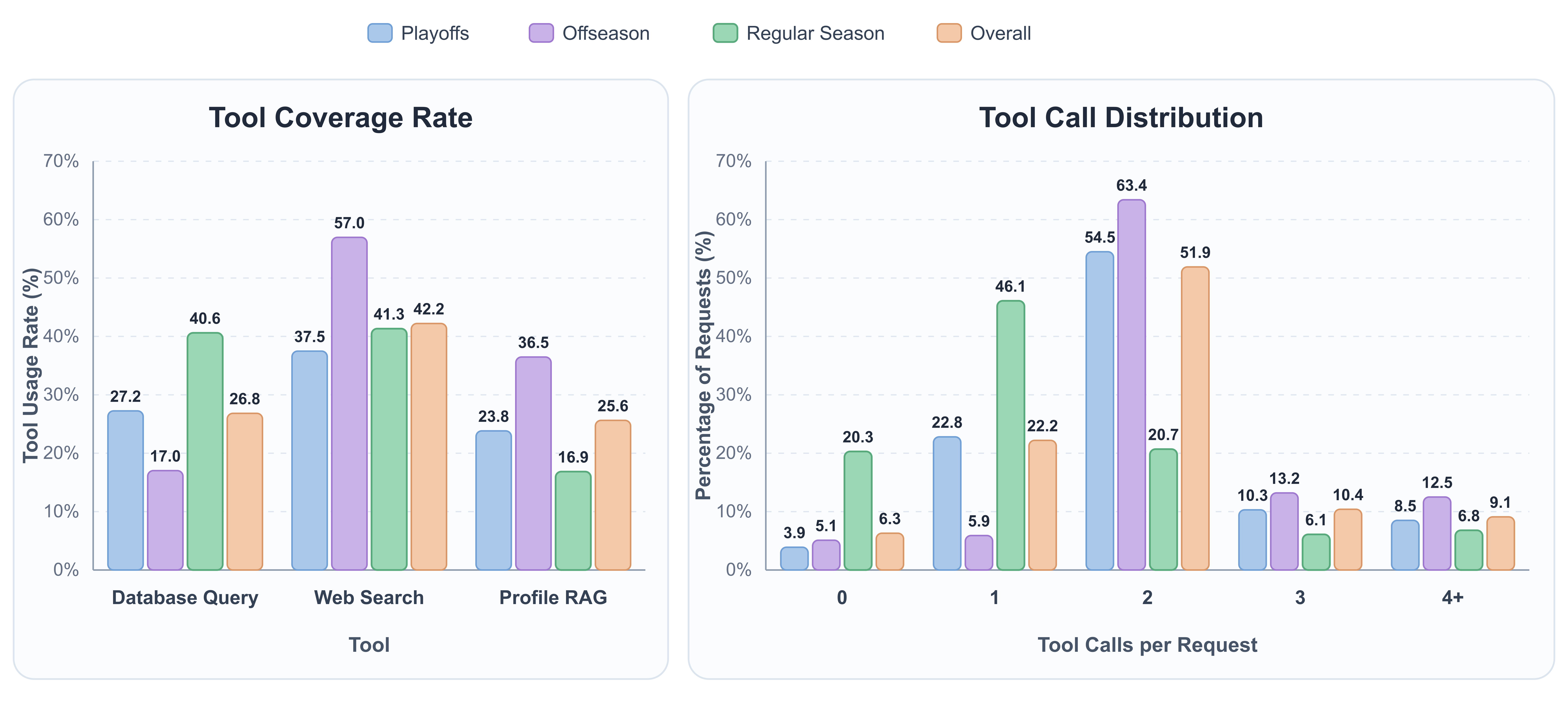}
  \caption{Tool usage and execution complexity across season phases. The left panel reports the coverage of database tool, web search, and profile RAG; the right panel shows the number of tool calls per request.}
  \label{fig:tool-usage-comparison}
\end{figure}

\subsection{User Engagement and Retention Comparison}

We compare user engagement, retention, and conversation depth over two 30-day observation periods, one in July 2026 for the baseline system and the other in October 2026 for the evolved system. We report five engagement-oriented metrics. \textbf{Interactions per User} measures the average number of user questions submitted per unique user. \textbf{Users with at Least Three Sessions} captures the proportion of users who initiate three or more distinct sessions during the observation window. \textbf{Users Active on at Least Three Days} measures repeated usage across different calendar days. \textbf{D7 Retention} is the proportion of eligible users who return exactly seven days after their first observed active date. Finally, \textbf{Sessions with at Least Three Turns} measures the proportion of reconstructed sessions containing three or more user questions, reflecting conversation depth and sustained follow-up behavior. Note that the absolute values are withheld to protect user privacy and commercially sensitive usage data.

\begin{figure}[htbp]
  \centering
  \includegraphics[width=\linewidth]{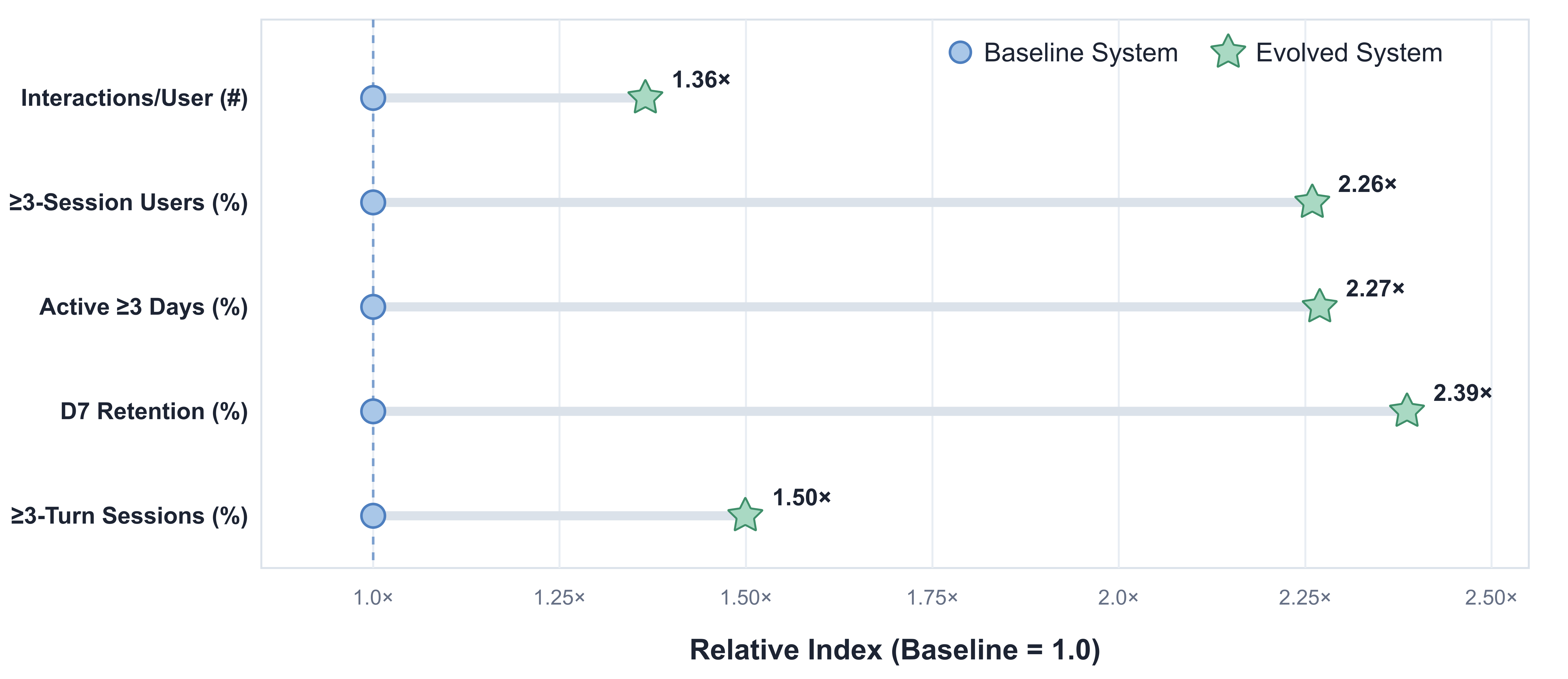}
  \caption{Relative user engagement, retention, and conversation depth for the evolved system, with the baseline normalized to 1.0 for each metric.}
  \label{fig:nba-chat-engagement}
\end{figure}

As shown in Figure~\ref{fig:nba-chat-engagement}, with the baseline system normalized to 1.0 for each metric, the evolved system shows improvements across all five engagement dimensions. Interactions per user reach \(1.36\times\) the baseline. The relative indices for users with at least three sessions and users active on at least three days increase to \(2.26\times\) and \(2.27\times\), respectively. D7 retention reaches \(2.39\times\) the baseline, while the share of sessions with at least three turns increases to \(1.50\times\). These results indicate greater conversation depth and repeated usage during the evolved system's observation window. Higher multi-session usage, cross-day activity, D7 retention, and multi-turn session rates suggest stronger overall user engagement. However, because the two evaluation windows cover different calendar periods and competition phases, the observed differences cannot be attributed solely to the version upgrade.

%% file: Sports_Agent_Anonymization/RelatedWork.tex
\section{Related Work}
\label{sec:related-work}

Sports Agent connects three research directions: harness evolution, model evolution, and agent diagnosis. This section reviews related work in these areas and discusses how the principles derived from our production deployment can inform the design of self-evolving agents in other domains, together with their applicability and limitations.

\subsection{Harness Self-Evolution}
\label{sec:related-harness-evolution}
\paragraph{From Prompt Optimization to Harness Evolution.}
Early adaptation methods focused on prompt optimization. \textbf{Promptbreeder}~\citep{fernando2024promptbreeder} jointly evolves task prompts and their mutation prompts, while \textbf{DSPy}~\citep{khattab2024dspy} compiles declarative LM modules against program-level metrics. \textbf{GEPA}~\citep{agrawal2026gepa} uses execution trajectories, including reasoning, tool calls, and outputs, to guide prompt updates. These methods establish prompts and modular inference programs as explicit optimization targets. Harness-oriented methods broaden the editable space to persistent context and executable structures. \textbf{ACE}~\citep{zhang2026agentic} maintains evolving strategy playbooks, while \textbf{AFlow}~\citep{zhang2025aflow} searches executable workflows composed of LLM-invoking nodes. \textbf{Automated Design of Agentic Systems}~\citep{hu2025adas} generates and archives agent programs through a meta-agent, and \textbf{STOP}~\citep{zelikman2024stop} recursively improves its own scaffolding under an explicit utility function. More directly, \textbf{Self-Harness}~\citep{zhang2026selfharness} groups verifier-grounded failures, proposes localized harness revisions, and selects candidates through regression gates. \textbf{HarnessEvolve}~\citep{jiang2026harnessevolve} uses validated reference trajectories to guide failure diagnosis, then screens updates through quality and performance gates. Together, these approaches extend adaptation from individual prompts to workflow structures and execution logic, with recent methods explicitly coupling failure diagnosis to validated harness updates ~\citep{lin2026agenticharness}.

\paragraph{Lifelong Learning of LLM Agents.}
Lifelong agent learning aims to retain and reuse execution experience across tasks. \textbf{Reflexion}~\citep{shinn2023reflexion} stores verbal feedback in episodic memory, while \textbf{Voyager}~\citep{wang2024voyager} accumulates executable skills through environment interaction. \textbf{ACE}~\citep{zhang2026agentic} maintains evolving strategy playbooks, and broader frameworks organize historical interactions into reusable memories and skills~\citep{cai2025lifelong}. \textbf{Living-Harness}~\citep{du2026livingharness} further converts trajectories and evaluator feedback into bounded updates of memory and control rules. These methods turn feedback into persistent capabilities, but retained artifacts may become outdated or redundant. More evolution rounds do not guarantee gains, making validation essential~\citep{liu2026rethinking}. Sports Agent aggregates related failures across dates and batches into typed evolution intents, proposing updates only when evidence is sufficient. Validation then selects effective candidates, enabling controlled adaptation while limiting incident-level overfitting.

\subsection{Model Self-Evolution}
\label{sec:related-model-evolution}

\paragraph{From Self-Training to Autonomous Post-Training.}
Model self-evolution improves successive checkpoints through generated supervision and iterative feedback. Bootstrapped methods such as \textbf{STaR} and \textbf{ReST$^{\mathrm{EM}}$} generate, filter, and learn from model outputs~\citep{zelikman2022star,singh2023restem}, while self-play and model-generated judgments support alternative training objectives~\citep{chen2024spin,yuan2024selfrewarding}. More recent work explores automation of the learning procedure itself: model-development benchmarks evaluate end-to-end experimentation~\citep{chan2025mlebench,nathani2025mlgym}, and research agents select data, training methods, and hyperparameters~\citep{yano2025lamdagent,rank2026posttrainbench}. This progression extends evolution from generating supervision to designing the training process. However, automated training may amplify model bias, exploit evaluation weaknesses, or introduce data contamination, making independent verification and data provenance essential.

\paragraph{Data and Trajectory Evolution.}
Recent work increasingly couples training-data generation with environment interaction and learner feedback. Environment-grounded task discovery and execution-verified self-play create supervision from agent interactions and executable problems~\citep{zhai2025agentevolver,zhao2025absolutezero}, while capability-adaptive generators produce tool-use tasks near the solver's current competence boundary~\citep{acikgoz2026toolr0}. Trajectory-based methods construct and filter teacher demonstrations or repair failed rollouts for subsequent training~\citep{wu2025webdancer,yuan2025agentr}. Failure-driven curricula further use diagnostic feedback to generate targeted tasks during reinforcement learning~\citep{wang2026sentinel}. These approaches automate supervision, but generated tasks still require verification, and targeted expansion must avoid overly narrow coverage. 

Sports Agent combines executable task synthesis with structured diagnosis of observed failures. SQL-to-Question Scaling converts verified, schema- and difficulty-controlled SQL into natural-language questions. Graph-Guided Scaling uses an attribution graph to target diagnosed capability gaps while varying entities, constraints, and table combinations. Together, these pathways use agent diagnosis to guide data generation and executable checks to validate the results.

\subsection{Agent Evaluation and Diagnosis}
\label{sec:related-agent-diagnosis}

\paragraph{From Outcome Scoring to Process-Aware Evaluation.}
LLM-based evaluators provide a scalable alternative to human review. \textbf{G-Eval}~\citep{liu2023geval} evaluates generated text using explicit criteria and structured reasoning, while \textbf{MT-Bench} and \textbf{Chatbot Arena}~\citep{zheng2023llmjudge} examine the reliability and bias of LLM judges. For factuality, \textbf{FActScore}~\citep{min2023factscore} decomposes responses into atomic claims and verifies them against external knowledge. Agent evaluation increasingly extends from final answers to intermediate processes. \textbf{AgentBoard}~\citep{ma2024agentboard} measures task progress in addition to final success. \textbf{Agent-as-a-Judge}~\citep{zhuge2025agentjudge} and \textbf{AgentDiagnose}~\citep{ou2025agentdiagnose} evaluate intermediate decisions and capabilities such as planning, observation reading, exploration, and self-verification. These methods show that outcome scores alone are insufficient for diagnosing tool-using agents; evaluation must retain the local evidence and trajectory segments that support each judgment.

\paragraph{Root-Cause Attribution in Agent Systems.}
A detected error does not directly identify the component that should be updated. An unsupported answer may result from missing model knowledge, an incomplete Prompt, failed Skill activation, or incorrect tool use. Failure detection and repair localization are therefore distinct problems. Recent work highlights the difficulty of reliable attribution. \textbf{MAST}~\citep{cemri2025mast} provides a taxonomy of failures in multi-agent systems, while \textbf{TRAIL}~\citep{deshpande2025trail} evaluates failure-category prediction and span-level localization over agent traces. Automated attribution methods also attempt to identify both the responsible agent and the decisive failure step~\citep{zhang2025whoandwhen}. These studies show that failure classification is easier than precise localization, particularly for long trajectories. Reliable diagnosis therefore requires trace decomposition and evidence-backed hypothesis testing rather than a single judgment over the complete execution.

Sports Agent separates quality assessment from root-cause diagnosis. For confirmed errors, the Feedback Engine replays tasks and inspects context, Prompts, Skills, and tool interactions to test competing explanations. Diagnoses specify causes, evidence, update targets, and actions, guiding model training, harness updates, regression testing, and release decisions. Diagnosis thus connects runtime evidence to controlled system evolution.

%% file: Sports_Agent_Anonymization/Discussion.tex
\section{Discussion: Broader Lessons for Self-Evolving Agents}
\label{sec:related-synthesis}

\paragraph{Positioning Sports Agent.}
Related research has expanded the scope of agent evolution from prompts to executable workflows, model training, and failure diagnosis. Recent methods already connect diagnosis with validated Harness revisions~\citep{zhang2026selfharness,jiang2026harnessevolve} and coordinate Harness adaptation with model updates~\citep{chen2026coharness,chen2026evotrainer,mao2026safeevolve}. Sports Agent builds on these advances by addressing three linked production questions: which failures justify persistent changes, which capabilities should be learned rather than patched, and how diagnostic findings become validated updates. Table~\ref{tab:related-work-positioning} summarizes this positioning.

\begin{table*}[t]
  \centering
  \caption{Positioning Sports Agent with respect to related research directions.}
  \label{tab:related-work-positioning}
  \small
  \setlength{\tabcolsep}{5pt}
  \renewcommand{\arraystretch}{1.2}
  \begin{tabularx}{\textwidth}{
      @{}
      >{\raggedright\arraybackslash}p{0.15\textwidth}
      >{\hsize=0.85\hsize\linewidth=\hsize\raggedright\arraybackslash}X
      >{\hsize=1.0\hsize\linewidth=\hsize\raggedright\arraybackslash}X
      >{\hsize=1.15\hsize\linewidth=\hsize\raggedright\arraybackslash}X
      @{}}
    \toprule
    \textbf{Direction}
      & \textbf{Optimization Object}
      & \textbf{Key Open Question}
      & \textbf{Key Mechanisms of Our Approach} \\
    \midrule[0.9pt]
    Harness evolution
      & Prompts, skills, context, tools, and workflows
      & Which failures justify updates, and how can rule accumulation be controlled?
      & Evidence accumulation guides updates; Validated consolidation controls redundancy. \\
    \arrayrulecolor{black!20}
    \midrule[0.3pt]
    \arrayrulecolor{black}
    Model self-evolution
      & Training data, model weights, and training strategies
      & How can generated supervision address deployment gaps while preserving validity and coverage?
      & Verified SQL anchors scalable question synthesis; Attribution-guided synthesis broadens coverage beyond observed failures. \\
    \arrayrulecolor{black!20}
    \midrule[0.3pt]
    \arrayrulecolor{black}
    Agent evaluation and diagnosis
      & Quality metrics and failure attribution
      & How can observed errors be translated into targeted, validated updates?
      & Evidence-grounded assessment checks correctness; failure attribution guides targeted updates. \\
    \bottomrule
  \end{tabularx}
\end{table*}

The contribution is a controlled workflow linking runtime evidence to persistent system updates, rather than a standalone optimization algorithm. Evidence support justifies proposals, diagnosis identifies update targets, validation governs acceptance, and consolidation limits complexity. 

\paragraph{Design Principles.}
Drawing on our system design and production experience, we summarize five principles for controlled agent evolution beyond the sports domain:

\begin{enumerate}
    \item \textbf{Attribute before updating.}
    Evidence-grounded diagnosis should guide model training, Harness revision, or both. Ambiguous attribution requires further verification rather than immediate modification.

    \item \textbf{Avoid updates based on isolated failures.}
    Individual failures may reflect ambiguous requests or stochastic generation. Persistent updates should therefore rely on sufficient evidence from related cases accumulated over time, rather than isolated incidents.

    \item \textbf{Generalize mechanisms, not individual failures.}
    Directly synthesizing variants of failed examples can cause local overfitting and evaluation leakage. Instead, identify the underlying capability gaps and use them to generate diverse training examples that extend beyond the original cases.

    \item \textbf{Ground supervision in verifiable representations.}
    SQL-to-Question synthesis illustrates a broader principle: use executable or machine-checkable representations to validate generated tasks and training signals, while retaining semantic checks on questions, trajectories, and answers.

    \item \textbf{Simplify the Harness as capabilities improve. }
    As the agent improves, review existing rules and skills and remove those that are no longer needed. Validate each simplification through regression testing and retain the option to restore the previous version.
\end{enumerate}

\paragraph{Cross-Domain Transferability.}
The proposed principles can extend to agents whose execution traces are observable, the components requiring updates can be identified, and outcomes can be independently assessed. Transfer requires adapting the diagnostic categories, evidence sources, verification methods, and release criteria to the target domain. For coding agents, test results and execution traces can support failure diagnosis, while repository state identifies the code or configuration to update. Enterprise assistants require checks that distinguish retrieval failures, outdated knowledge, and unsupported answers. Multimodal agents additionally need evidence alignment across text, images, and video, including temporal consistency. These adaptations connect the shared evolution process to each domain’s failure mechanisms and quality requirements.

%% file: Sports_Agent_Anonymization/Conclusion.tex
\section{Summary and Outlook}
\label{sec:conclusion}

This work investigated how a production sports agent can sustain reliable conversational data analytics under continuous changes in user demand, data resources, and model behavior. We presented Sports Agent, a dual-loop self-evolution framework that connects online execution with offline system improvement through a shared Feedback Engine. The framework organizes runtime queries, responses and execution trajectories into structured evidence; diagnoses whether an observed failure originates from the model or the harness; and converts the diagnosis into actionable evolution signals. The model loop expands training coverage through SQL-to-Question scaling and graph-guided synthesis, then uses generated trajectories for supervised fine-tuning. A separate Post-RL stage further refines the resulting model. In parallel, the harness loop applies evidence-supported updates to prompts, skills, tool policies, and response logic, while dual audits, regression replay, and non-inferiority guards control update risk and system complexity.
Experiments demonstrate that the proposed approach improves both capability and execution efficiency. On the Sports Agentic SQL generation task, the evolved model improves accuracy from 83.5\% to 97.8\% on the Sports-Normal test set and from 45.7\% to 74.0\% on the Sports-Hard test set, while substantially reducing latency and tool-call rounds. The gains extend to general Agentic SQL benchmarks, supporting the transferability of the learned query and tool-use capabilities beyond the evaluated sports databases. In the deployed product, the evolved system increases the factual correctness score from 75.8 to 86.3 and average response length by 158.1\%, while reducing average end-to-end latency by 41.3\%. Across 30 successive harness-evolution rounds, the aggregate Performance Score rises from 78.2 to 89.0. Together, these results show that higher answer quality does not necessarily require longer execution paths: targeted model training and more effective orchestration can improve reliability and efficiency simultaneously.

The results demonstrate that runtime evidence can improve model capabilities and agent orchestration. However, model and harness updates remain largely separate, harness evolution primarily targets prompts and skills, and the current workflow connects evidence-based evaluation, diagnosis, and service updates while still requiring substantial human involvement in code review and production deployment. These limitations motivate three directions for future work.

\textbf{Model--Harness Co-Evolution.}
Building on recent co-evolution frameworks~\citep{chen2026evotrainer,mao2026safeevolve,chen2026coharness}, future work will investigate alternating optimization: execution feedback guides harness revisions, and quality-filtered trajectories generated under the revised harness support model training. Joint credit assignment will help determine whether failures require model updates, harness revisions, or both. Shared quality and efficiency objectives will coordinate these updates, reducing component mismatch and redundant modifications.

\textbf{Agent Design and Workflow Evolution.}
Future work will extend harness evolution beyond prompts and skills to executable workflows and orchestration code. Modularizing workflow stages and atomic operations will enable search over agent structures and implementations~\citep{zhang2025aflow,wang2026harness}. Execution evidence can guide changes to workflow topology, control logic, and individual code blocks. Constrained generation, sandboxed execution, static analysis, and regression testing will support reliable structural updates.

\textbf{Full-Lifecycle Agent Self-Evolution.}
While evidence-based evaluation, failure diagnosis, and service updates are already automated, code review, deployment, and release verification remain manual. Future work will close this loop, linking failure discovery to candidate implementation, automated review, deployment, and post-release validation, with every change tied to a versioned artifact and traceable evidence of its effects.

%% file: Sports_Agent_Anonymization/Appendix.tex
\appendix
\section{Experimental Setup for the Agentic SQL Model}
\label{app:agentic-sql-setup}

\subsection{Benchmarks and Evaluation Settings}

\paragraph{Sports Agentic SQL.}
We construct a domain-specific benchmark for production sports question answering and analysis, comprising Sports-Normal and Sports-Hard. Sports-Normal contains 200 questions derived from real-world requirements, covering player and team information, schedules and results, basic statistics, and conditional filtering. It evaluates routine query capabilities and execution efficiency. Sports-Hard contains 100 challenging query tasks involving multi-table joins, compound constraints, cross-season comparisons, multi-stage aggregation, and complex game-context analysis. All questions are manually reviewed for domain relevance and answerability using the available databases. Together, the two subsets assess performance on routine queries and complex, long-tail analytical tasks.

\paragraph{LiveSQLBench.}
LiveSQLBench evaluates agentic text-to-SQL on real databases, covering both SELECT and management/CRUD tasks. We evaluate all three released splits with the official LiveSQLBench-Agent ADK harness.\footnote{\url{https://github.com/bird-bench/livesqlbench/tree/main/LiveSQLBench-Agent}} The agent receives only the query and a 30-tool-call budget; it retrieves schema and external knowledge through tools, can execute SELECT/WITH/EXPLAIN queries to inspect intermediate results, and submits exactly once. We report the official reward: 1 if the submitted SQL passes all test cases and 0 otherwise (i.e., the task success rate).

\paragraph{BIRD-Interact.}
BIRD-Interact evaluates interactive text-to-SQL under ambiguity, where the agent must clarify with a user simulator and explore the database. We evaluate both splits in a-Interact mode with the official BIRD-Interact-ADK harness.\footnote{\url{https://github.com/bird-bench/BIRD-Interact/tree/main/BIRD-Interact-ADK}} The agent has nine actions with bird-coin costs and a stress-mode interaction budget; Qwen3.8-Flash is used as the user simulator. We report the normalized reward (NR), denoted Reward in Table~\ref{tab:agentic-sql-benchmark-performance}; in a-Interact, NR is 1.0 when both sub-tasks pass, 0.7 when only the first sub-task passes, and 0 otherwise.

For both public benchmarks, our evaluation follows the official ADK implementations: the agent scaffold, prompts, and budget are identical to the official code, and the only difference is the evaluated model endpoint (and the BIRD-Interact user simulator). All runs use temperature 0.0 and a single run per task. Tool-use efficiency is measured by the mean number of tool-call rounds per task.

\begin{table}[htbp]
  \centering
  \caption{Evaluation data used in Agentic SQL experiments.}
  \label{tab:agentic-sql-evaluation-data}
  \begin{tabular}{lc}
    \toprule
    Benchmark & Size \\
    \midrule
    \multicolumn{2}{l}{\textbf{Sports Agentic SQL}} \\
    Sports-Normal & 200 \\
    Sports-Hard & 100 \\
    \midrule
    \multicolumn{2}{l}{\textbf{General Agentic SQL}} \\
    LiveSQLBench-Base-Lite & 270 \\
    LiveSQLBench-Base-Full & 600 \\
    LiveSQLBench-Large & 480 \\
    BIRD-Interact-Lite & 300 \\
    BIRD-Interact-Full & 600 \\
    \bottomrule
  \end{tabular}
\end{table}

\subsection{Post-RL Training Details}
\label{app:agentic-sql-rl}
Table~\ref{tab:rl-reward-values} summarizes the instantiated rewards. The complete definitions and implementation details follow.

\begin{table}[htbp]
  \centering
  \caption{Instantiated reward values for the two task types.}
  \label{tab:rl-reward-values}
  \linespread{1.10}\selectfont          % 单元格内部折行行距
\renewcommand{\arraystretch}{1.35}    % 表格行与行之间的留白
  \small
  \begin{tabular}{p{0.20\linewidth}p{0.60\linewidth}p{0.14\linewidth}}
    \toprule
    Task Type & Scoring Rule & Raw Reward \\
    \midrule
    Execution-verifiable & Predicted and reference queries have equivalent results; staged tasks accumulate stage rewards & 0, 0.7, 1 \\
    Open-ended analysis & Repeated LLM-judge decisions & 0, 1/3, 2/3, 1 \\
    \bottomrule
  \end{tabular}
\end{table}

\paragraph{Objective.}
Post-RL adopts GSPO~\citep{gspo}. For each prompt, the system samples a group of trajectories and normalizes rewards within the group to obtain advantage $A_i$. The sequence-level importance ratio, computed only over model-generated tokens and excluding tool observations, is
\begin{equation}
s_i(\theta) = \left( \frac{\pi_\theta(\tau_i \mid x)}{\pi_{\theta_{\mathrm{old}}}(\tau_i \mid x)} \right)^{1/|\tau_i|}.
\end{equation}
Training maximizes
\begin{equation}
\mathcal{J}(\theta) = \mathbb{E}\left[ \frac{1}{G} \sum_{i=1}^{G} \min\left( s_i(\theta) A_i,\ \mathrm{clip}\left(s_i(\theta), 1-\epsilon, 1+\epsilon\right) A_i \right) \right],
\end{equation}
where $G$ is the group size, $|\tau_i|$ is the number of model-generated tokens in the trajectory, and $\epsilon$ is the clipping coefficient.

\paragraph{Execution-Verifiable Tasks.}
Single- and multi-stage tasks use a common stage-weighted execution-equivalence reward:
\begin{equation}
R_{\mathrm{exec}} = \sum_{s=1}^{C_{\mathrm{exec}}} w_s \cdot \mathbb{1}\left[\mathrm{exec}(\hat{y}_s) \equiv \mathrm{exec}(y_s^*)\right],
\qquad
w_s =
\begin{cases}
1, & C=1,\\
0.7, & C=2,\ s=1,\\
0.3, & C=2,\ s=2.
\end{cases}
\end{equation}
$R_{\mathrm{exec}}$ is the execution-equivalence reward, $s$ is the stage index, $w_s$ is its weight, $C\in\{1,2\}$ is the number of stages defined by the task, and $C_{\mathrm{exec}}\in\{1,\ldots,C\}$ is the number of executable stages. $\hat{y}_s$ and $y_s^*$ are the predicted and reference queries. The maximum reward is
\begin{equation}
\max R_{\mathrm{exec}} =
\begin{cases}
1, & C=1,\\
0.7, & C=2,\ C_{\mathrm{exec}}=1,\\
1, & C=2,\ C_{\mathrm{exec}}=2.
\end{cases}
\end{equation}
A single-stage task receives binary execution-equivalence credit. For a two-stage task, the first successful stage contributes 0.7 and the second contributes 0.3. If the task has no executable second stage, the maximum is 0.7. Stages are evaluated in order, and a later stage contributes only after the preceding stage succeeds.

Before comparison, queries are normalized by removing comments and selected non-semantic operators such as DISTINCT and ROUND. Results are compared as sets unless the task explicitly requires ordering. Empty results receive zero. Each trajectory may submit only once; SQL errors, missing submissions, empty or inequivalent results, and failed execution all receive zero without partial credit. Retries within a stage are not penalized. When the interaction budget is exhausted, the server permits one final submission and terminates in the next round, preventing repeated free submissions.

\paragraph{Open-Ended Analysis Tasks.}
These tasks have no unique reference SQL and are scored by an LLM judge. Each trajectory is judged independently $k=3$ times, and its reward is the resolved fraction:
\begin{equation}
R_{\mathrm{judge}} = \frac{1}{k} \sum_{i=1}^{k} \mathbb{1}[\text{resolved}_i].
\end{equation}
We use Qwen3.7-Max as the judge and provide the input query together with the temporally ordered SQL--observation pairs. The judge assigns structured 0--5 ratings for intent coverage, query validity, and trajectory efficiency and returns a resolved decision. Missing key constraints, adding or removing filters that change the result, substituting an approximate metric, answering only part of the question, or entering a loop yields false. Boundary-expression differences such as $>$ versus $\geq$ do not. The continuous \texttt{resolved\_rate} is used directly; repeated evaluations at the same temperature reduce service and parsing variance rather than forming a majority vote. A trajectory is sent to the judge only after at least one tool call and a valid completion action. Judge failures or unparseable outputs receive zero, and the judge evaluates the query trajectory rather than the final natural-language response.

\paragraph{Sampling Strategy.}
Rollout uses dynamic sampling~\citep{yu2026dapo}: prompt groups with near-zero reward variance---all correct or all incorrect---are discarded and resampled, ensuring that every group used for gradient updates has non-zero advantage. The procedure is implemented during rollout assembly without changing the main training loop.

\subsection{Tool-Use Efficiency on General Agentic SQL}
For General Agentic SQL, efficiency is reported as the mean number of tool-call rounds; latency is not used.

\begin{table}[htbp]
  \centering
  \caption{Average tool-call rounds on General Agentic SQL.}
  \label{tab:general-tool-call-rounds}
  \begin{tabular}{lcc}
    \toprule
    & \multicolumn{2}{c}{Average Tool-Call Rounds} \\
    \cmidrule(lr){2-3}
    Model & LiveSQLBench & BIRD-Interact \\
    \midrule
    Qwen3.6-35B-A3B & 15.8 & 20.8 \\
    Qwen3.8-Max & 14.4 & 17.6 \\
    Ours & 9.4 & 11.3 \\
    \bottomrule
  \end{tabular}
\end{table}

Our model averages 9.4 tool-call rounds on LiveSQLBench and 11.3 on BIRD-Interact, reductions of 34.7\% and 35.8\% relative to Qwen3.8-Max. The evolved model therefore solves tasks on both general benchmarks with substantially fewer interactions.

\section{Privacy and Data Protection}
\label{app:data-privacy}

The analyses presented in this report use only de-identified, task-level records required for system evaluation, including input queries, model responses, and task-relevant agent execution trajectories. These records are processed without user labels, persistent identifiers, or account-level metadata. No direct user identifiers - such as names, phone numbers, email addresses, IP addresses, or account identifiers, or profile and preference attributes are collected in this study. Workload and engagement statistics are reported only in aggregate, and no attempt is made to identify, profile, or link records to individual users. Access to the analysis data is strictly limited to system evaluation and improvement.